\documentclass[runningheads]{llncs}

\usepackage{eccv}

\usepackage{eccvabbrv}
\usepackage{orcidlink}
\usepackage{graphicx}
\usepackage{booktabs}

\usepackage[accsupp]{axessibility}  % Improves PDF readability for those with disabilities.

\usepackage{multirow}
\usepackage{graphicx}
\usepackage{algorithm}
\usepackage{algpseudocode}
\usepackage{array}
\newcolumntype{C}[1]{>{\centering\arraybackslash}p{#1}}
\usepackage{hyperref}

\usepackage{orcidlink}
\usepackage{soul}
\usepackage{amsmath}
\DeclareMathOperator*{\argmin}{arg\,min} 

\begin{document}

% ---------------------------------------------------------------
% TODO REVIEW: Replace with your title
\title{Flatness-aware Sequential Learning Generates Resilient Backdoors} 

% TODO REVIEW: If the paper title is too long for the running head, you can set
% an abbreviated paper title here. If not, comment out.
% \titlerunning{Sequential Backdoor Learning}

% TODO FINAL: Replace with your author list. 
% Include the authors' OCRID for the camera-ready version, if at all possible.
\author{Hoang Pham\inst{1}\orcidlink{0009-0004-2501-8271} \and
The-Anh Ta\inst{2}\orcidlink{0000-0003-2615-7316} \and
Anh Tran\inst{3}\orcidlink{0000-0002-3120-4036} \and
Khoa D. Doan\inst{1}\orcidlink{0000-0002-1610-8206}
}

% TODO FINAL: Replace with an abbreviated list of authors.
\authorrunning{Hoang Pham et al.}
% First names are abbreviated in the running head.
% If there are more than two authors, 'et al.' is used.

% TODO FINAL: Replace with your institution list.
\institute{College of Engineering and Computer Science, VinUniversity \\
\and
CSIRO's Data61 \quad\quad\quad  \inst{3} VinAI Research \\
\email{hoang.pv1602@gmail.com}, \quad \email{khoa.dd@vinuni.edu.vn} \\
\email{theanh.ta@csiro.au}, \quad \email{v.anhtt152@vinai.io}}

\maketitle

\begin{abstract}
Recently, backdoor attacks have become an emerging threat to the security of machine learning models. From the adversary's perspective, the implanted backdoors should be resistant to defensive algorithms, but some recently proposed fine-tuning defenses can remove these backdoors with notable efficacy. This is mainly due to the catastrophic forgetting (CF) property of deep neural networks. This paper counters CF of backdoors by leveraging continual learning (CL) techniques. We begin by investigating the connectivity between a backdoored and fine-tuned model in the loss landscape. Our analysis confirms that fine-tuning defenses, especially the more advanced ones, can easily push a poisoned model out of the backdoor regions, making it forget all about the backdoors. Based on this finding, we re-formulate backdoor training through the lens of CL and propose a novel framework, named \textbf{S}equential \textbf{B}ackdoor \textbf{L}earning (\textbf{SBL}), that can generate resilient backdoors. This framework separates the backdoor poisoning process into two tasks: the first task learns a backdoored model, while the second task, based on the CL principles, moves it to a backdoored region resistant to fine-tuning. We additionally propose to seek flatter backdoor regions via a sharpness-aware minimizer in the framework, further strengthening the durability of the implanted backdoor. Finally, we demonstrate the effectiveness of our method through extensive empirical experiments on several benchmark datasets in the backdoor domain. The source code is available at \url{https://github.com/mail-research/SBL-resilient-backdoors}

% Recently, backdoor attacks have become an emerging threat to the security of machine learning models. From the adversary's perspective, the implanted backdoors should be resistant to defensive algorithms, but some recently proposed fine-tuning defenses can remove these backdoors with notable efficacy. This paper studies a backdoor insertion framework capable of circumventing these fine-tuning defenses. We begin by investigating the connectivity between a backdoored and fine-tuned model in the loss landscape. Our analysis confirms that fine-tuning defenses, especially the more advanced ones, can easily push a poisoned model out of the backdoor regions, making it forget all about the backdoors. Based on this finding, we re-formulate backdoor training through the lens of continual learning (CL) and propose a novel framework, named \textbf{S}equential \textbf{B}ackdoor \textbf{L}earning (\textbf{SBL}), that can leverage CL techniques to generate resilient backdoors. This framework separates the backdoor poisoning process into two tasks: the first task learns a backdoored model, while the second task, based on the CL principles, moves it to a backdoored region resistant to fine-tuning. We additionally propose to seek flatter backdoor regions via a sharpness-aware minimizer in the framework, further strengthening the durability of the implanted backdoor. Finally, we demonstrate the effectiveness of our method through extensive empirical experiments on several benchmark datasets in the backdoor domain.
\keywords{Backdoor attack \and Continual learning }
\end{abstract}

\vspace{-0.5cm}
\section{Introduction}
\label{sec:intro}

% {\color{orange} texts need to paraphrase}

Swift progress in the field of machine learning (ML), especially within the realm of deep neural networks (DNNs), is revolutionizing various aspects of our daily lives across different domains and applications from computer vision to natural language processing tasks~\cite{radford2019language, radford2021learning, rombach2022high, brown2020language, wei2022finetuned}.
Unfortunately, as well-trained models are now considered valuable assets due to the significant computational resources, annotated data, and expertise spent to create them, they are becoming appealing targets for cyber attacks~\cite{liu2018survey, carlini2021extracting, carlini2017towards}.
Prior studies have shown that DNN models are vulnerable to diverse attacks, from exploratory attacks such as adversarial attacks~\cite{yuan2019adversarial, kumar2020adversarial, qin2019imperceptible} to causative attacks such as poisoning attacks~\cite{guo2020practical, shafahi2018poison} and backdoor attacks~\cite{gu2017badnets,  nguyen2021wanet, nguyen2024backdoor}. 
Among these, backdoor attacks have recently gained attention because of the increasing popularity of machine learning as a service (MLaaS), where a model user outsources model training to a more experienced ML service provider.  
In backdoor attacks, the adversary injects a backdoor into the poisoned model by either contaminating the training data~\cite{gu2017badnets, chen2017targeted, doan2022marksman, nguyen2020input, nguyen2021wanet, barni2019new} or manipulating the training process~\cite{zhang2022how, feng2022stealthy, garg2020can, doan2021lira, doan2021back}. This backdoored model is expected to behave normally on benign input, but give a specific output, defined by the attacker, when the backdoor trigger appears on any input. 
Consequently, the attacker can deceive the model user into integrating this poisoned model (with the hidden backdoor) into their systems to gain illegal benefits or cause harmful damages~\cite{li2022backdoor, goldblum2022dataset,zhang2022backdoor,iba}.

\begin{figure}[ht]
    \centering
    \includegraphics[width=1\linewidth]{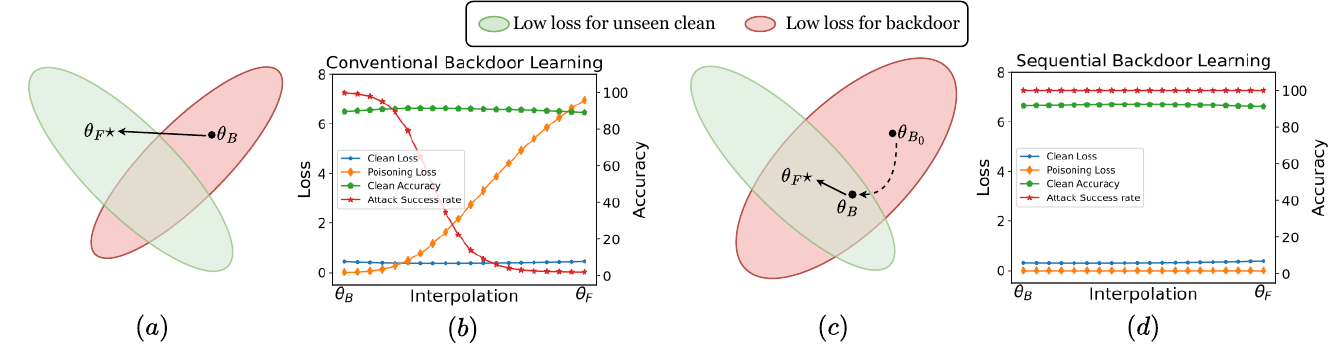}
    \vspace{-0.3cm}
    \caption{(a) Intuition for fine-tuning defense against conventional backdoor learning (CBL): the backdoored model $\theta_B$ is pushed out of backdoor region (red area);
    (c) Intuition for the success of our sequential backdoor learning (SBL) framework: $\theta_{B}$ is trapped within the backdoor region that is hard to escape with fine-tuning.
    %in which models locate in the area of low-loss low backdoor learning loss? for backdoor perform well on both clean and poisoned data.
    %'Backdoor region' refers to region of models with good performance on both poisoned and clean data - the red area.
    Figure b and d visualize the loss and the accuracy on clean and poisoned test sets of intermediate models when linearly interpolating between backdoored and fine-tuned models with CBL and SBL.}
    \label{fig:intuition}
    \vspace{-0.5cm}
\end{figure}

% \begin{figure}
%   \centering

%   % First row with one figure
%   \begin{subfigure}{\linewidth}
%     \centering
%     \includegraphics[width=\linewidth]{figures/CBL_intuition.pdf}
%     \caption{Comparison between normal backdoor learning (left) and sequentially backdoor learning (right)}
%     \label{fig:intuition}
%   \end{subfigure}
  
%   % Second row with two figures
%   \begin{subfigure}{0.48\linewidth}
%     \centering
%     \includegraphics[width=\linewidth]{figures/Normal_BL.pdf}
%     \caption{}
%     \label{fig:nbl_interpolation}
%   \end{subfigure}
%   \hfill
%   \begin{subfigure}{0.48\linewidth}
%     \centering
%     \includegraphics[width=\linewidth]{figures/CBL.pdf}
%     \caption{}
%     \label{fig:cbl_interpolation}
%   \end{subfigure}

%   \caption{The loss and the accuracy on clean and poisoned test sets of intermediate models when linearly interpolating between backdoored and fine-tuned models with normal backdoor learning (b) and sequentially backdoor learning (c).}
% \end{figure}

As research on backdoor attacks advances, numerous defense strategies against such attacks have been introduced. These defenses can detect the poisoned model~\cite{wang2019neural, gao2019strip}, or remove the backdoors by knowledge distillation~\cite{li2021neural, huang2023distilling}, and pruning~\cite{wu2021adversarial, liu2018fine}. 
While being effective, these methods pose utility (e.g., non-trivial drop in accuracy), and consistency (e.g., effectiveness dependent on network architectures) challenges~\cite{Zhu_2023_ICCV}. 
% While being effective, these methods pose usability challenges, particularly for individuals who are not experts in the field~\cite{Zhu_2023_ICCV}. 
% In contrast, fine-tuning~\cite{liu2018fine, guo2019spottune, wei2022finetuned} is a more general approach not only for backdoor defense but also for many practical AI training pipelines when users utilize pre-trained models. 
Recently, fine-tuning defenses~\cite{liu2018fine, Zhu_2023_ICCV} have shown promising performance in backdoor removals; furthermore, since the model user usually does not use a pre-trained model as-is but will adapt it for their problem, fine-tuning is likely a necessary step in many practical ML pipelines. In particular, the user can fine-tune a pre-trained model using a small clean dataset~\cite{liu2018fine, Zhu_2023_ICCV}, making it forget the implanted backdoor. 
% as visualized in Figure~\ref{fig:intuition} (top-left). 
As demonstrated in Figure~(\ref{fig:intuition}$b$),
%the top right of Figure~\ref{fig:intuition},
when we linearly interpolate from the backdoored model (with conventional backdoor learning) to its corresponding fine-tuned model, the intermediate model's poisoning loss (i.e., the loss recorded on the poisoned samples only) increases, resulting in the decrease in Attack Success Rate (ASR) accordingly. Meanwhile, the corresponding clean loss and accuracy are stable. This indicates that fine-tuning with clean data can move the poisoned model to the clean-only region, explaining the effectiveness of existing fine-tuning defenses. 
% Besides, [{\color{blue}{add citations}}] have leveraged fine-tuning to achieve better defense performances. 
% \hl{[TheAnh: The above paragraph is too long, perhaps list out the points you want to make and the logic, then rewrite it into a more concise paragraph]}

We take the perspective of the backdoor attackers and aim to investigate backdoors that are resilient to fine-tuning based defenses. 
Our starting point is the observation that the main reason for the effectiveness of fine-tuning defense is  the \textit{catastrophic forgetting} property of DNNs \cite{li2022backdoor} when models are continuously trained on unseen clean data. 
The setting is then naturally connected to Continual Learning (CL) - the learning paradigm that focuses exactly on mitigating catastrophic forgetting \cite{kirkpatrick2017overcoming, aljundi2018memory, chaudhry2018efficient}.
Thus, to counter the use of forgetting to cleanse backdoors in fine-tuning defenses, our high-level idea is to leverage CL techniques to craft backdoors that are hard to forget.

Specifically, we design a novel sequential learning procedure for backdoor attack and propose a new framework, named SBL, for creating fine-tuning resistant backdoored models. 
% We view backdoor attack and defense as tasks in CL setting, the goal of attackers is crafting backdoored models avoiding backdoor forgetting after defense. In CL, when learning new task, the model immediately adapts to the new data and forget knowledge acquired in prior tasks. This is the mechanism of fine-tuning based defense methods when the backdoored model is continuously trained on clean-only data. We counteract this phenomenon by making the backdoored model familiarize itself with clean data beforehand.
% Specifically, our framework 
SBL separates the backdoor learning process into two sequential tasks: the first task learns the backdoored model, while the second task simulates fine-tuning defense on this model with a small set of clean data. We augment the second task with CL techniques to guide the poisoned model towards a low-loss, backdoored minimum (from $\theta_{B_0}$ to $\theta_B$ in Figure~(\ref{fig:intuition}$c)$) where it is difficult for the defender to remove the backdoor with fine-tuning defenses. 
% for clean data but still be inside the backdoored area. 
% low-loss minima for clean data but still be inside the backdoored area (from $\theta_{B_0}$ to $\theta_B$ in Figure~\ref{fig:intuition}). 

We additionally seek for \textit{flat} backdoor region that can intensify the backdoor eliminating challenge for a fine-tuning defense, further strengthening the durability of the implanted backdoor. 
Our goal is to trap the model in a flat backdoored area that is hard to escape. 
% This sequential learning framework, augmented with CL techniques and flatness-aware learning, essentially traps the learned backdoored model in a backdoor region that is hard to escape with existing fine-tuning based defenses. 
% \hl{[theanh: the previous sentence and the following sentence are not linked, can you connect these two sentences to have a logical paragraph?]}
We demonstrate the effectiveness of our SBL in Figure~(\ref{fig:intuition}$d)$ revealing that as we interpolate linearly between the backdoored and the fine-tuned models, the poisoning loss and ASR remain largely constant. Meanwhile, the clean performance experiences a slight improvement along the connectivity path.

In summary, our main contributions are:
\vspace{-0.1cm}
\begin{itemize}
    \item[(i)] We propose a novel backdoor learning framework, named SBL, which involves two sequential learning tasks, for generating resistant backdoored models. The framework is inspired by the empirical observation that existing fine-tuning defenses can effectively push the backdoored model to a backdoor-free region in the parameter space. Our learning framework can be used to train existing backdoor attacks, further improving their resistance against fine-tuning defenses.

    \item[(ii)] We propose to formulate the second task from the continual learning perspective. This involves applications of existing CL techniques and a flatness-aware minimization approach, both of which collaboratively strengthen the resistance of the backdoor.

    \item[(iii)] We perform extensive empirical experiments and analysis to demonstrate the effectiveness of the proposed framework in improving the durability of several existing backdoor attacks. This urges backdoor researchers to devise defensive measures to counter this type of attack.
\end{itemize}

\vspace{-0.2cm}
\section{Related Work}
\label{sec:related_work}

% \hl{TheANh:what are previous works on backdoor attacks that aim to dodge the fine-tuning defense? if there are none then we can claim that this is the first work focus on fine-tuning defense, all other attacks are easily mitigated by fine-tuning (perhaps can add to introduction to highlight). I mean we need to compare this work to previous works, not just list out the related works.}

% \hl{theanh: the related work section seems too long}

\subsection{Backdoor Attacks} 
In backdoor attacks, the adversary aims to manipulate the output of the victim model to a specific target label with input having pre-defined triggers~\cite{gao2020backdoor, dai2019backdoor, li2020backdoor, ramakrishnan2020backdoors}. The backdoor injection process can be done by poisoning data ~\cite{chen2017targeted, gu2017badnets} or maliciously implanting a backdoor during training~\cite{zhang2022how, garg2020can}. 
Gu et al~\cite{gu2017badnets} first investigated backdoor attacks in deep neural networks and proposed BadNets. It injects the trigger into a small random number of inputs in the training set and re-labels them into target labels. After that various backdoor attacks focus on designing the triggers. In particular, Chen et al~\cite{chen2017targeted} leverage image blending in design trigger while Barni et al~\cite{barni2019new} use sinusoidal strips. WaNet~\cite{nguyen2020input} trains a generator to create input-aware triggers. LIRA~\cite{doan2021lira,doan2021back} jointly learns trigger generator and victim model to launch imperceptible backdoor attacks. Besides data attacks, some works~\cite{zhang2022how, garg2020can} perturb the weights of a pre-trained clean model to inject a backdoor.

\subsection{Backdoor Defenses}
% {\color{orange}The victim could be aware of or advised about the security threats at every stage of building and utilizing the model, thus they could apply defenses in all stages, ranging from data scanning (data defense) and model examination (model defense) to test-time monitoring after the model is deployed (test-time defense). Our paper focuses on fine-tuning-based defenses, a popular and straightforward approach in model defense. }

In general, backdoor defense methods can be divided into pre-training~\cite{wang2019neural, tran2018spectral}, in-training~\cite{zhang2023backdoor, li2021anti}, and post-training~\cite{liu2018fine, li2021neural, huang2023distilling, Zhu_2023_ICCV, liu2019abs} stages. In pre-training and in-training defenses, the defender assumes the dataset is poisoned, and thus leverages the models' distinct behavioral differences on the clean and poisoned samples to remove the manipulated data or avoid learning the backdoor during training. Most defensive solutions perform post-training defenses since it can be more challenging to alter or, in some cases, not possible to participate in the training process. Post-training defenses assume the defender has access to a small set of benign samples for backdoor removal~\cite{liu2018fine, li2021neural, Zhu_2023_ICCV, hu2022trigger, wu2021adversarial} and can be roughly categorized into fine-tuning based defenses~\cite{liu2018fine, li2021neural, Zhu_2023_ICCV} and pruning-based defenses~\cite{liu2018fine, wu2021adversarial, chai2022one}. 
Pruning-based defenses prune neurons~\cite{liu2018fine, wu2021adversarial} or weights~\cite{chai2022one} to remove backdoor-contaminated components in the model. However, these methods either cause non-trivial drops in benign accuracy or their effectiveness depends on the network architectures~\cite{Zhu_2023_ICCV}, significantly reducing their utility and consistency.  
On the other hand, fine-tuning based defenses leverage the catastrophic forgetting phenomenon of DNNs~\cite{kirkpatrick2017overcoming, aljundi2018memory}, when a backdoored model is fine-tuned on clean data, for backdoor removal. In addition, fine-tuning is also a common step in numerous practical ML systems to adapt a pre-trained model to better align with the user's needs. 
This paper focuses on developing a novel backdoor learning approach that can enable existing backdoor attacks to be resistant to conventional fine-tuning processes in practical applications and advanced fine-tuning defenses. 
% , a widely used and straightforward technique in both model defense and practical applications.

\subsection{Continual Learning}
Continual Learning (CL) is a learning paradigm where the model learns a sequence of tasks. When tasks arrive, the model has to preserve previous knowledge while efficiently learning new tasks, which is known as the stability-plasticity dilemma. There are three main approaches to dealing with this problem. 
\begin{itemize}
    \item[(i)] \textbf{Regularization} approaches~\cite{kirkpatrick2017overcoming, zenke2017continual, aljundi2018memory, van2022auxiliary, v.2018variational, ahn2019uncertainty} add explicit regularization terms to penalize the variation of each parameter using its ``importance" in performing the old tasks. 
    \item[(ii)] \textbf{Architecture-based} approaches~\cite{yoon2018lifelong, yan2021dynamically, mallya2018packnet, mallya2018piggyback, pmlr-v162-gurbuz22a} assign specific parameters for each task and even expand the base architecture when more parameters are required. 
    \item[(iii)] \textbf{Replay-based} approaches~\cite{lopez2017gradient, chaudhry2018efficient, bang2021rainbow, deng2021flattening, prabhu2020gdumb} store a set of prior-task data and use the stored data together with new-task data when learning on new tasks. 
\end{itemize}

Our method views backdoor attack and defense as a continual learning problem, where backdoor learning is the first task, and fine-tuning on clean, unseen data is another task. 

\subsection{Mode Connectivity and Sharpness-Aware Minimization}
Loss landscape has been investigated to understand the behavior of DNNs~\cite{li2018visualizing, garipov2018loss, foret2021sharpnessaware}. Hochreiter et al~\cite{hochreiter1997flat} show that flat and wide minima generalize better than sharp minima. Recently, SAM~\cite{foret2021sharpnessaware} and its variants~\cite{kwon2021asam, liu2022towards, zhang2023gradient} improve generalization by simultaneously minimizing both the loss value and loss sharpness.
This property is leveraged to mitigate forgetting in CL methods~\cite{mirzadeh2020understanding, deng2021flattening}.
Besides, Mode Connectivity~\cite{draxler2018essentially, garipov2018loss}, a novel tool to understand the loss landscape, postulates that different optima obtained by gradient-based optimization methods are connected by simple low-error path (i.e., low-loss valleys). 
% \cite{neyshabur2020being} investigated the connection between minima obtained by pre-trained models versus freshly initialized ones. They note that there is no performance barrier between solutions coming from pre-trained models, but there can be a barrier between solutions of different randomly initialized models. 
% \cite{frankle2020linear} shows that different minima that share the same initialization point are connected by a linear path, even with weight pruning. 
Mirzadeh et al~\cite{mirzadeh2021linear} observe that there exists a low-error path connecting multi-task and continual learning minima when they share a common starting point. Motivated by this observation, the works in~\cite{lin2022towards, mirzadeh2021linear} propose methods to guide the model towards this connectivity region. 

\section{Methodology}
\label{sec:method}

\subsection{Threat Model}

% paraphrase this paragraph

We adopt the commonly-used backdoor-attack setting where the attacker trains a model and provides it to the victim~\cite{li2022backdoor,doan2021back}. Since training large-scale neural networks is empirical, data-driven, and resource-extensive, it is generally cost-prohibitive for end-users, who consequently turn to third-party MLaaS platforms~\cite{ribeiro2015mlaas} for model training, or simply clone pre-trained models from public sources such as Hugging Face. This practice opens up opportunities for training-control backdoor attacks, a serious security threat to victim users. 
% This is a practical assumption because training large-scale neural networks is empirical, data-driven, and resource-extensive, making it unaffordable for end-users who consequently turn to third-party services like MLaaS~\cite{ribeiro2015mlaas} platforms, or simply clone pre-trained models from public sources such as Hugging Face. 
%
% Such common practices create venues for attackers and make backdoor attacks a serious security threat to victim users. 

\vspace{0.2cm}
\noindent \textbf{Attacker's Capability.} The attacker has full control of designing the triggers, poisoning training data, and the model training schedule.

\vspace{0.2cm}
\noindent \textbf{Attacker's Goal.} The attacker aims to implant a backdoor into the model and bypass post-training defense methods, especially fine-tuning defenses.
%such that it can not be eliminated when users do fine-tuning on a small set of clean data or apply defense methods. 

% \vspace{-0.5cm}
\noindent \textbf{Defender's Goal.} Focusing on the recent fine-tuning defenses, we assume that the victim is given a pre-trained backdoored model, and has access to a small, clean dataset. 
The defender's goal is to then fine-tune the model on the clean data to remove potentially hidden backdoors while adapting and maintaining the model's performance on their data.

% \theanh{Perhaps rewrite the whole section 3.1. Give exact and concise thread model for standard backdoor attack - this is well established now, what is the capacity of the attacker in this standard model and its goals, finetuning defense. Can be formal and precise, no need MLaaS or any application here, or maybe just mention it quickly, as I think it is mentioned in the introduction or related works already.}

\subsection{Conventional Backdoor Learning}
\label{sec:conventional_backdoor_training}

We consider supervised learning of classification tasks where the objective is to learn a mapping function $f_\theta: \mathcal{X} \rightarrow \mathcal{C}$ with the input domain $\mathcal{X}$ and the set of class labels $\mathcal{C}$. 
The task is to learn the parameters $\theta$ from training dataset $\mathcal{D} = \{(x_i, y_i)  : x_i \in \mathcal{X}, y_i \in \mathcal{C}, i = 1, 2, ..., N \}$ using a standard classification loss $\mathcal{L}$ such as Cross-Entropy Loss. The most common training scheme for backdoor attacks uses data poisoning to implant backdoors, where the classifier is trained on $\mathcal{D}_p$ - a mixture of clean and poisoned data from $\mathcal{D}$. 
The general procedure to generate poisoned data is to transform a clean training sample $(x, y)$ into a backdoor sample $(T (x), \eta (y))$ with some backdoor injection function $T$ and target label function $\eta$. 
Backdoor training manipulates the behavior of $f$ so that: $f(x) = y, \, f(T(x)) =  \eta (y)$.
%
%\begin{align*}
%    f(x) = y, \quad f(T(x)) =  \eta (y).
%\end{align*}
%
%backdoor region
%region of models with good performance on both poisoned and clean data
% In the traditional paradigm of backdoor learning, the model is jointly trained on a combination of clean and poisoned data to inject backdoors. 

It is well-established that such training can cause the models to converge to the backdoor regions. 
However, empirical evidence \cite{liu2018fine} (see also our Figure \ref{fig:loss_analysis}) suggests that even a simple fine-tuning process with a small set of clean data can lead the model to an alternative local minimum that is free of backdoor while preserving the model's performance on clean data.
%Fine-tuning have been one of the most effective methods for backdoor defenses.
% Our analysis reveals that after convergence, the model's loss landscape becomes rugged for both backdoor and clean data losses, indicating that minor perturbations to the model's parameters can result in the backdoor elimination.

\subsection{Proposed SBL Framework}
\label{sec:proposed_framework}

This paper views backdoor learning through the lens of continual learning (CL): we re-formulate the attack and defense as the CL tasks.
%as in the CL paradigm.
% The backdoored model is created by attackers, then defenders clean this model by fine-tuning on clean data.
More precisely, the attacker aims to develop resilient backdoors that remain even after the models undergo fine-tuning defenses at the user's site - this can be regarded as reducing catastrophic forgetting in CL; 
while the defender strives to relocate the models away from the backdoor region without compromising performance on clean data - leveraging catastrophic forgetting to remove backdoors.

To challenge the effectiveness of fine-tuning defenses, our key idea is to simulate this defense mechanism during the training phase of backdoor learning to familiarize our models with clean-data fine-tuning, which reduces the effect of forgetting during any subsequent fine-tuning defenses.
%
%As adversaries, we counter these strategies by simulating defense mechanisms during the backdoor learning phase. 
%
In particular, we split the training data $\mathcal{D}_p$ into two sets $\mathcal{D}_0$ and $\mathcal{D}_1$, where $\mathcal{D}_0$ is a combination of clean and poisoned data while $\mathcal{D}_1$ contains only clean samples. Then, we divide backdoor training into two consecutive tasks: first to learn the backdoor, then familiarize it with fine-tuning. In the first step, we learn a backdoored model $\theta_{B_0}$ on $\mathcal{D}_0$ by utilizing Sharpness-Aware Minimization (SAM)~\cite{foret2021sharpnessaware,kwon2021asam} on the loss $\mathcal{L}^{SAM}(\mathcal{D}_0; \theta)$. Since a flatter loss landscape is known to reduce catastrophic forgetting~\cite{mirzadeh2020understanding, deng2021flattening}, this training strategy will seek a flat backdoor loss landscape, consequently limiting the model's ability to forget backdoor-related knowledge during fine-tuning defenses with clean data. 
% When learning the second task, we augment training with CL principles to guide this backdoored model toward a region with low clean loss but deeper within the backdoor's effective area. 
In the second step, we continue to train $\theta_{B_0}$ (found in Step 0) on $\mathcal{D}_1$ with relatively \textit{small learning rate} and the additional CL regularization to force the model to converge into a low clean loss basin 
% with the backdoored model $\theta_{B_0}$ 
but deeper within the backdoor's effective area:
% Additionally, since a flatter loss landscape is known to reduce catastrophic forgetting~\cite{mirzadeh2020understanding, deng2021flattening}, we utilize Sharpness-Aware Minimization (SAM)~\cite{foret2021sharpnessaware, kwon2021asam}, when training the poisoned model, to seek for a flat backdoor loss landscape, consequently limiting the model's ability to forget backdoor-related knowledge during fine-tuning defenses with clean data.
% Specifically, we split the training data $\mathcal{D}_p$ into two sets $\mathcal{D}_0$ and $\mathcal{D}_1$, where $\mathcal{D}_0$ is a combination of clean and poisoned data while $\mathcal{D}_1$ contains only clean samples. In the first step, we learn a backdoored model $\theta_{B_0}$ on $\mathcal{D}_0$ by minimizing the loss $\mathcal{L}^{SAM}(\mathcal{D}_0; \theta)$, where SAM~\cite{foret2021sharpnessaware} is utilized to guide the model converging toward flat backdoor region. 
%
% In the second step, we continue to train $\theta_{B_0}$ on $\mathcal{D}_1$ with relatively \textit{small learning rate} and regularization to force the model to converge into the same loss basin with the backdoored model $\theta_{B_0}$: 
%
\begin{align}
\small
    % \mathcal{L}_1 = \mathcal{L}(\mathcal{D}_1; \theta_{B_0}) + \mathcal{R}(\theta)
     \mathcal{L}_1 = \mathcal{L}(\mathcal{D}_1; \theta) + \mathcal{R}(\theta_{B_0}, \theta)
    % \vspace{-2cm}
\end{align}

\vspace{-0.7cm}
\begin{algorithm}
\caption{Sequential Backdoor Learning (SBL)}
\label{alg:sbl_framework}
\begin{algorithmic}[1]
\State \textbf{Input:} Training data $\mathcal{D}_0$, $\mathcal{D}_1$, model's parameters $\theta$

\State \textbf{Output:} Backdoored model $\theta_{B}$

\State Initialize model parameter $\theta$

\State \textbf{Step 0:} Learning the first task
\State \quad $\theta_{B_0} \leftarrow \argmin\limits_{\theta} \, \mathcal{L}^{SAM}(\mathcal{D}_0; \theta)$

\State \textbf{Step 1:} Fine-tuning on clean data with constrains
\State\quad Set $\theta \leftarrow \theta_{B_0}$
\State \quad $\theta_{B} \leftarrow \argmin\limits_{\theta} 
\mathcal{L}(\mathcal{D}_1; \theta) + \mathcal{R}(\theta_{B_0}, \theta)$

% \texttt{minimize} \,  \mathcal{L}(\mathcal{D}_1; \theta_{B_0}) + \mathcal{R}(\theta)$

\State \textbf{Return:} $\theta_B$

\end{algorithmic}
\end{algorithm}

\vspace{-0.7cm}

% \subsection{Why should our method work in practice?}
\subsection{On the working mechanism of our method SBL} \label{sec:why_sbl_works}

%While our method SBL looks simple and is easy to implement, experiments (Section \ref{sec:experiments}) show that it is surprisingly effective in training backdoor models resilient  against powerful fine-tuning defense methods. 
% To better understand the effectiveness of SBL, we discuss here motivations and heuristic explanations for its working mechanism.
%In particular, we seek to answer two questions: $(i)$ What properties that models are expected to attain under SBL training? and $(ii)$ Why should SBL be able to produce such models?

The proposed method SBL is designed based on our intuition that to neutralize the effect of fine-tuning defenses, we can train the model so that it converges to backdoored regions having flat loss landscapes. 
Flatness then can cause the fine-tuned model in Step 1 (usually with small learning rates) to still be trapped in the region of backdoor knowledge, which makes our attack resilient to fine-tuning defenses. 
Here, we provide further heuristic explanations based on observations from continual learning and mode connectivity, and confirm with empirical evidence that the behaviors of our approach closely align with these intuitions. 
Additional analysis in terms of Taylor expansions is given in the Appendix.

We regard the Algorithm \ref{alg:sbl_framework} of SBL as a two-step procedure: multi-task (MT) training (\textbf{Step 0}) followed by continual learning (CL) (\textbf{Step 1}). 
More precisely,
SBL first trains the backdoored model $\theta_{B_0}$ on both clean and poisoned data (MT), then fine-tunes $\theta_{B_0}$ with clean data and a tiny learning rate to obtain $\theta_B$ (CL). 
Denote $\theta_{F}$ the model obtained with a fine-tuning defense afterward. 

\begin{figure}[ht]
  \centering
  % Second row with two figures
  \raisebox{5mm}{\rotatebox{90}{\tiny ResNet18 on CIFAR-10}}%
  \begin{subfigure}{0.32\linewidth}
    \centering
    \includegraphics[width=\linewidth]{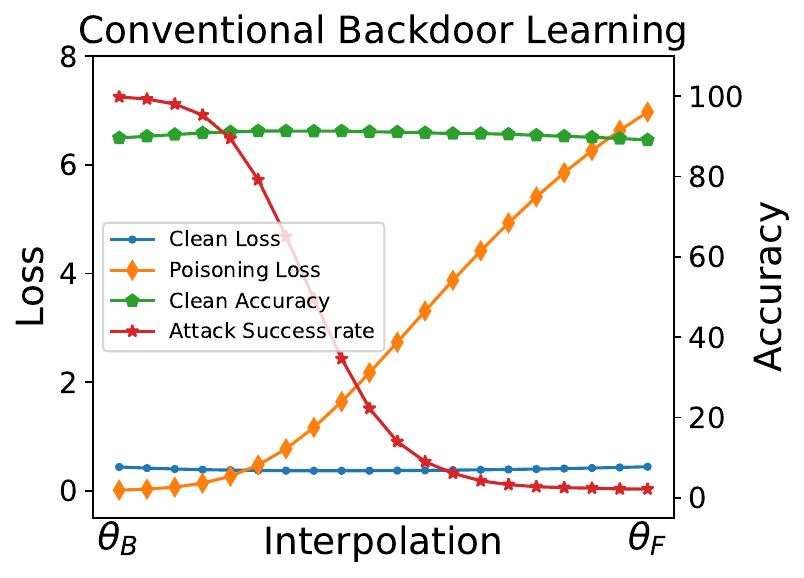}
    \label{fig:MI_cifar_badnet_joint}
  \end{subfigure}
\hfill
  \begin{subfigure}{0.32\linewidth}
    \centering
    \includegraphics[width=\linewidth]{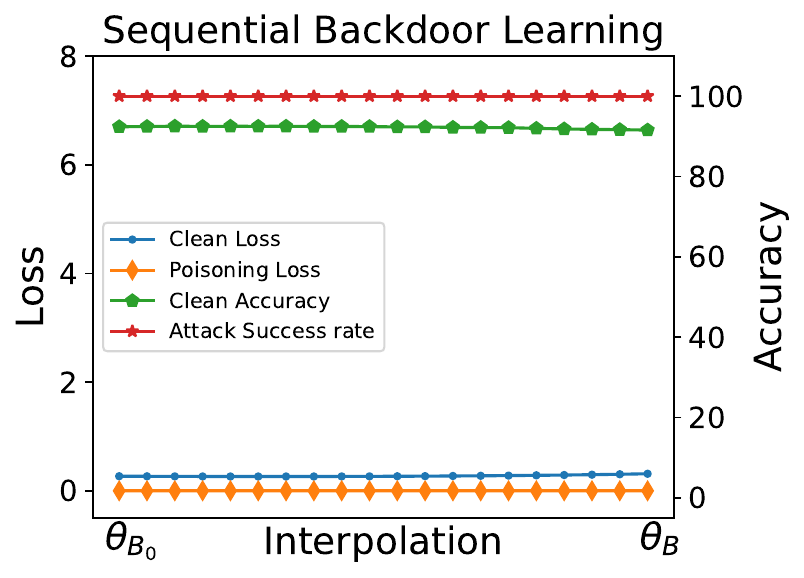}
    \label{fig:MI_cifar_badnet_sbl_f2c}
  \end{subfigure}
  \hfill
  \begin{subfigure}{0.32\linewidth}
    \centering
    \includegraphics[width=\linewidth]{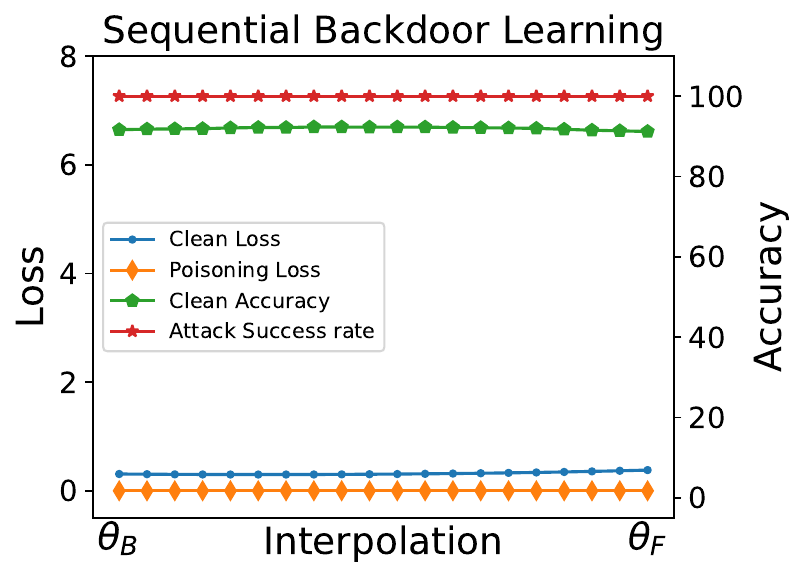}
    \label{fig:MI_cifar_badnet_sbl}
  \end{subfigure}

\vspace{-0.4cm}
% Second row with two figures
    \raisebox{6mm}{\rotatebox{90}{\tiny ResNet18 on GTSRB}}%
  \begin{subfigure}{0.32\linewidth}
    \centering
    \includegraphics[width=\linewidth]{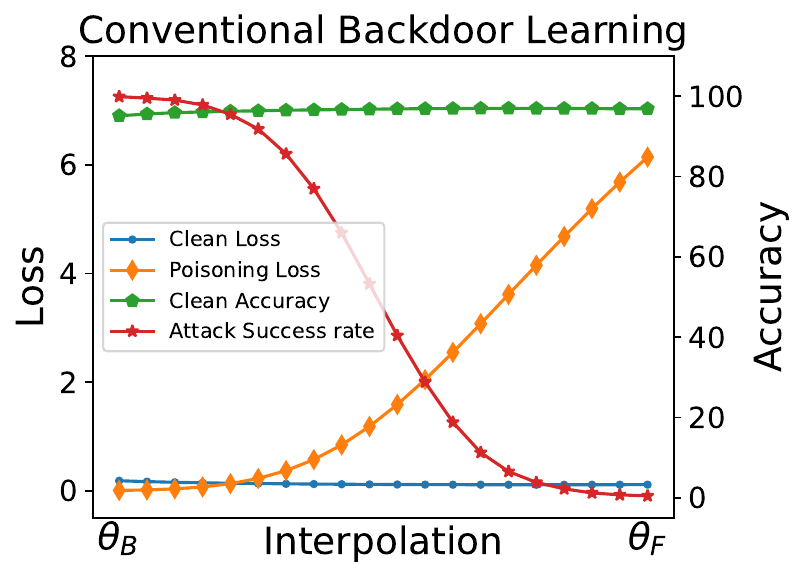}
    \label{fig:MI_gtsrb_badnet_joint}
  \end{subfigure}
\hfill
  \begin{subfigure}{0.32\linewidth}
    \centering
    \includegraphics[width=\linewidth]{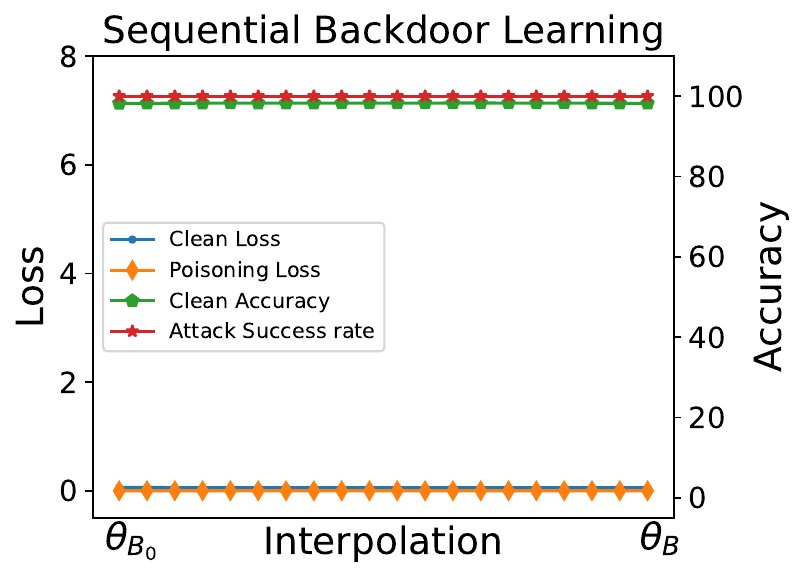}
    \label{fig:MI_gtsrb_badnet_sbl_f2c}
  \end{subfigure}
  \hfill
  \begin{subfigure}{0.32\linewidth}
    \centering
    \includegraphics[width=\linewidth]{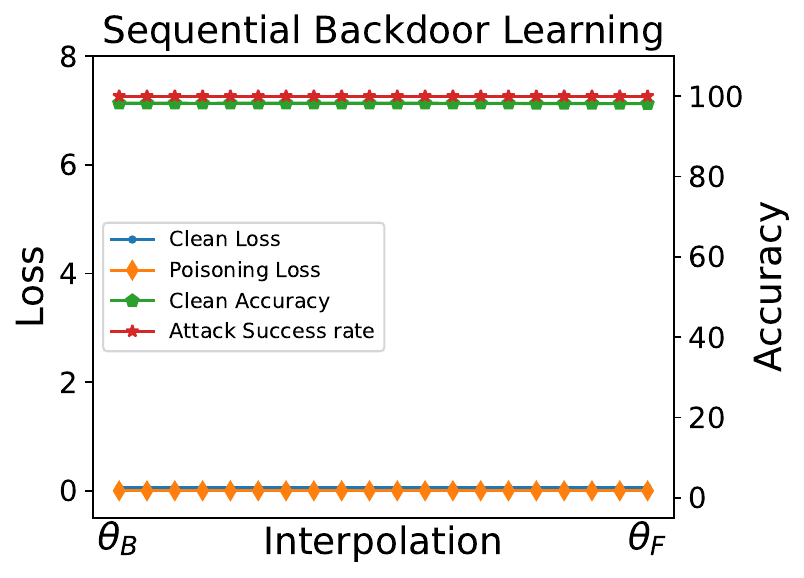}
    \label{fig:MI_gtsrb_badnet_sbl}
  \end{subfigure}
  
  \vspace{-0.4cm}

  \caption{The loss and the accuracy on clean and poisoned test sets of intermediate models when linearly interpolating between models. The first column is between backdoored and fine-tuned models in conventional backdoor learning, the second column is between models in the first ($\theta_{B_0}$) and second task ($\theta_B$), while the last column is between backdoored and fine-tuned models in our SBL framework.}
  \label{fig:loss_analysis}
  \vspace{-0.5cm}
\end{figure}

First, we calculate the losses and accuracies of models trained with SBL along various connectivity paths and compare them to those obtained with conventional backdoor learning (CBL) models. We perform experiments on two settings: ResNet18 model on CIFAR-10 and GTSRB, with BadNets as the base attack. In Figure~\ref{fig:loss_analysis}, the first column visualizes the loss and accuracy on the clean and poisoned test sets, where we linearly interpolate between a backdoored and fine-tuned model in the CBL setup. While the clean loss and accuracy remain unchanged, the poison loss gradually increases and the corresponding ASR decreases to nearly zero on the fine-tuned model. This indicates that with CBL, fine-tuning can effectively push the poisoned model out of the backdoor-affected area. On the other hand, the persistently high ASR and low poison losses in the third column of Figure~\ref{fig:loss_analysis} (interpolations between $\theta_{B}$ and $\theta_{F}$) show that our SBL method can trap backdoored model in backdoored region that is difficult to escape from.

Recent works~\cite{mirzadeh2021linear, lin2022towards} have established that there are low-error pathways connecting minima of MT solutions and CL solutions.
In SBL, \textbf{Step 1} is designed to seek a low-error path to guide our model to a flat backdoored solution.
Empirically, we observe in Figure~\ref{fig:loss_analysis} that SBL can identify \textit{low-error pathways} connecting multi-task model ($\theta_{B_0}$) to continual learning model ($\theta_{B}$) (the second column), and from $\theta_{B}$ to the fine-tuned model ($\theta_{F}$) (the third column).
%\hl{Heuristically, our framework guides the backdoored model to follow piecewise approximations by a tiny learning rate of these low-loss paths, making it hard to escape from these regions in the later fine-tuning based defenses. -- this part is not clear} 
%Heuristically, in Step 1 our framework guides the backdoored model $\theta_{B_0}$ (MT solution) to follow these low-loss paths on finding low-loss minima for clean data $D_1$ by using a tiny learning rate and CL regularization. This makes the model $\theta_B$ (CL solution) hard to escape from these regions (low loss landscape for both clean and poisoned data) in the later fine-tuning based defenses.
% We provide empirical evidence to support this point of view through model interpolation between $\theta_{B_0}$ and $\theta_B$, $\theta_F$ in Figure \ref{fig:loss_analysis}. 
In addition, we empirically show in Figure~\ref{fig:grad_norm} that gradients' norm during fine-tuning defense on clean data remains small, which directly mitigates the forgetting of backdoor knowledge.

\vspace{-0.5cm}

\section{Experiments}
\label{sec:experiments}

\subsection{Experiment Setting}

\noindent \textbf{Datasets.} We use three benchmark datasets in backdoor research, namely CIFAR-10~\cite{krizhevsky2009learning}, GTSRB~\cite{Houben-IJCNN-2013}, and ImageNet-10 for the experiments. We follow \cite{huang2022learning} to select 10 classes from ImageNet-1K~\cite{imagenet15russakovsky}. 
We divide the training set into three subsets: mixed set $\mathcal{D}_0$ (poisoned and benign samples), clean set $\mathcal{D}_1$, and defense set with portion 85\% - 10\% - 5\%, respectively.

\vspace{0.3cm}
\noindent \textbf{Models.} We use the same classifier backbone ResNet18~\cite{he2016deep} for all datasets. We also employ different architectures (in ablation studies): VGG-16~\cite{simonyan2014very} and ResNet-20, a lightweight version of ResNet. We use SGD optimizer for training the backdoored model and SAM~\cite{foret2021sharpnessaware} for training the first task. We set the learning rate to $0.01$ in the first task and $0.001$ in the second task. We train the backdoored model for 150 epochs in Step 0 and for 100 epochs in Step 1. 
% Additional training details are provided in the Appendix.

\vspace{0.3cm}
\noindent \textbf{Backdoor Attacks.} We consider representative backdoor attacks, including BadNets~\cite{gu2017badnets}, Blended~\cite{chen2017targeted}, SIG~\cite{barni2019new}, and Dynamic~\cite{nguyen2020input}. For BadNets, we use a random-color, $3 \times 3$ square at the bottom right corner as the backdoor trigger on all datasets. We use pre-trained generators from \cite{nguyen2020input} to generate poisoned images for Dynamic Attack. 
In all experiments, we poison 10\% of training data and set the targeted class to label 0.

% \textbf{BadNets}~\cite{gu2017badnets} This is the first and most basic backdoor attack, in which the backdoor pattern is just a hand-picked image patch. We use a random color $3 \times 3$ square at the bottom-right corner as the backdoor trigger on all datasets.

% \textbf{Blended}~\cite{chen2017targeted} This attack chooses an outer image, e.g., a Hello Kitty image, and blends it with clean samples to create backdoor data.

% \textbf{SIG}~\cite{barni2019new} This attack uses sinusoidal strips as a backdoor trigger. 
\vspace{0.3cm}
\noindent \textbf{Defense Methods.} We evaluate the persistence of the backdoored models against fine-tuning-based defenses, including standard finetuning~\cite{liu2018fine} and the advanced finetuning approaches, SAM-FT~\cite{Zhu_2023_ICCV} and NAD~\cite{li2021neural}. For standard finetuning, we use SGD with two learning rates $0.01$ and $0.005$; similarly, for SAM-FT, we set the learning rate to $0.005$. We fine-tune these models for 50 epochs. For NAD, we fine-tune the teacher and student using a learning rate of $0.01$ for 20 epochs.

\vspace{0.3cm}
\noindent \textbf{Backdoor Training Methods.} Here, along with original backdoor training, we select several CL techniques to incorporate into our SBL framework to train backdoored models including Naive, EWC~\cite{kirkpatrick2017overcoming}, Anchoring~\cite{zhang2022how}, and AGEM~\cite{chaudhry2018efficient}. 

% \textbf{Original}: This is a conventional backdoor training. Particularly, we incorporate mixed dataset $\mathcal{D}_0$ and clean data $\mathcal{D}_1$ into one to train the backdoored model. 

% \textbf{Naive}: Learning new task without CL techniques.

% \textbf{EWC}~\cite{kirkpatrick2017overcoming}: This is a regularization CL method that identifies important weights via Fisher Information Matrix (FIM). 
% EWC penalizes changes in crucial weights, the regularization term is: $F (\theta - \theta_{B_0})$ in which $F$ is the FIM.

% \textbf{Anchoring}~\cite{zhang2022how}: This method is used in~\cite{zhang2022how} to force the model to output similar logits on clean data. It adds regularization term: $\sum_i^C (s_i(x; \theta) - s_i(x; \theta_{B_0}))$ where $C$ is set of classes, $x$ is benign sample, and $s_i(.)$ is the logit for class $i$.

% \textbf{AGEM}~\cite{chaudhry2018efficient}: This is a memory-based CL method that stores a buffer of data from the first task. In updating the model on the second task, the gradient update is projected in a direction that does not hamper the update of previous tasks. In particular, let $g$ be the gradient computed with the incoming mini-batch and $g_{ref}$ be the gradient computed with the same size mini-batch randomly selected from the memory buffer. 
% In A-GEM, if $g^{\top}g_{ref} \geq 0$, $g$  is used for gradient update but when $g^{\top}g_{ref} < 0$, $g$ is projected such that $g^{\top} g_{ref} = 0$. The gradient after projection is: $\tilde{g} = g - \frac{g^{\top} g_{ref}}{g_{ref}^{\top} g_{ref}} g_{ref}$
\vspace{0.3cm}
\noindent \textbf{Evaluation Metrics.} We use two common metrics to evaluate the performance of the backdoored models: Clean Accuracy (\textbf{CA}) to measure the performance on benign samples and Attack Success Rate (\textbf{ASR}) to measure the proportion of backdoor samples that are successfully misclassified to the targeted label.

\vspace{0.3cm}
\noindent Additional experimental details are provided in the Appendix.

% \begin{figure}[htb]
%     \centering % <-- added
% \begin{subfigure}{0.325\textwidth}
%   \includegraphics[width=\linewidth]{figures/interpolation/cifar10_resnet18_joint2defense.pdf}
%   \caption{image1}
%   \label{fig:1}
% \end{subfigure}\hfil % <-- added
% \begin{subfigure}{0.325\textwidth}
%   \includegraphics[width=\linewidth]{figures/interpolation/cifar10_resnet18_first2sec.pdf}
%   \caption{image2}
%   \label{fig:2}
% \end{subfigure}\hfil % <-- added
% \begin{subfigure}{0.325\textwidth}
%   \includegraphics[width=\linewidth]{figures/interpolation/cifar10_resnet18_sec2defense.pdf}
%   \caption{image3}
%   \label{fig:3}
% \end{subfigure}

% \medskip
% \begin{subfigure}{0.325\textwidth}
%   \includegraphics[width=\linewidth]{figures/interpolation/gtsrb_resnet18_joint2defense.pdf}
%   \caption{image4}
%   \label{fig:4}
% \end{subfigure}\hfil % <-- added
% \begin{subfigure}{0.325\textwidth}
%   \includegraphics[width=\linewidth]{figures/interpolation/gtsrb_resnet18_first2sec.pdf}
%   \caption{image5}
%   \label{fig:5}
% \end{subfigure}\hfil % <-- added
% \begin{subfigure}{0.325\textwidth}
%   \includegraphics[width=\linewidth]{figures/interpolation/gtsrb_resnet18_sec2defense.pdf}
%   \caption{image6}
%   \label{fig:6}
% \end{subfigure}
% \caption{Fasi del processo di impregnazione}
% \label{fig:images}
% \end{figure}

%%%%%%%%%%%%%%%%%%%%%%

%%%%%%%%%%%%%%%%%%%%%%%

\vspace{-0.3cm}
\subsection{Backdoor Performance}
\label{sec:main_results}

% \hl{this section should be revised. It's not very clear. Suggest adding the breakout for results with different CL techniques. We should also mention the CL techniques used in the method sec too, and explain why they should work. Currently, readers have no idea what CL techniques are and why they help the method.}

We verify the effectiveness of SBL by comparing it against the standard backdoor training using different fine-tuning defenses. We conduct the experiments on three datasets, including CIFAR-10, GTSRB, and Imagenet-10 with 10\% poisoning rate, and with ResNet18. 
In Step 1, we train the backdoored model $\theta_{B_0}$ obtained in Step 0 with different CL methods.
The results are presented in Table~\ref{tab:main_cifar10}, ~\ref{tab:main_gtsrb}, and~\ref{tab:main_imagenet10}, respectively. Due to the space limit, we present the SBL's experiments with Anchoring in the Appendix.

%%%%%%%%%
In the CIFAR-10 and GTSRB settings,
for conventional backdoor training, fine-tuning with a lower learning rate (0.005) can better preserve the model's utility but fails to effectively mitigate the backdoor. In contrast, fine-tuning with a larger learning rate (0.01) can effectively eliminate backdoors but comes at the cost of sacrificing clean-data performance.
Both FT-SAM and NAD can maintain high clean accuracy while effectively mitigating backdoors. 
%%%%%%%%%
On the other hand, SBL significantly enhances the backdoor's resilience against all the previously mentioned defensive methods. With different CL techniques, SBL effectively circumvents these defenses. 
% Additionally, SBL also contributes to an improvement in clean accuracy.
%%%%%%

In the case of ImageNet-10, although fine-tuning defenses with learning rates in our setting successfully eliminate the effect of backdoors in Original baselines, it has to sacrifice the model's utility. Learning backdoored model via SBL can significantly improve the resistance against these fine-tuning defenses, most of them still achieve at least 60\% ASR while the CA is almost above 70\%. We attribute this improvement to the effect of flat minima discovered by our SBL.

% These results demonstrate the effectiveness of our SBL framework in creating backdoored models that are resistant to fine-tuning based defenses.

%%%%%%%%%%%%%%%%%%%%%%%%%%%%%%%%%%%%%%%%%%%%%
%%%%%%%%%% RESNET-18 ON CIFAR-10 %%%%%%%%%%%%%%%
%%%%%%%%%%%%%%%%%%%%%%%%%%%%%%%%%%%%%%%%%%%%%

%%%%%%%%%%%%%%%%%%%%%%%%%%%%%%%%%%%%%%%%%%%%%
%%%%%%%%%% RESNET-18 ON CIFAR-10 %%%%%%%%%%%%%%%
%%%%%%%%%%%%%%%%%%%%%%%%%%%%%%%%%%%%%%%%%%%%%

% Please add the following required packages to your document preamble:
% \usepackage{multirow}
\vspace{-0.3cm}
\begin{table*}[]
\caption{The resilience against fine-tuning defenses in setting  ResNet18 on CIFAR-10.}
\label{tab:main_cifar10}
\resizebox{0.98\textwidth}{!}{
\small 
\centering

\begin{tabular}{c|l|cc|cc|cc|cc|cc|cc}

\toprule
\multirow{2}{*}{Attack} & \multicolumn{1}{c|}{\multirow{2}{*}{Training}} & \multicolumn{2}{c|}{Step 0}                     & \multicolumn{2}{c|}{Step 1}      & \multicolumn{2}{c|}{FT SGD-0.005} & \multicolumn{2}{c|}{FT SGD-0.01} & \multicolumn{2}{c|}{NAD}         & \multicolumn{2}{c}{FT-SAM} \\ \cline{3-14} 
                                  & \multicolumn{1}{c|}{}                                 & CA                   & \multicolumn{1}{c|}{ASR} & CA    & \multicolumn{1}{c|}{ASR} & CA         & \multicolumn{1}{c|}{ASR}     & CA        & \multicolumn{1}{c|}{ASR}     & CA    & \multicolumn{1}{c|}{ASR} & CA               & ASR             \\ \hline
\multirow{5}{*}{\textbf{Badnets}} & Original                                              & \multicolumn{1}{l}{} & \multicolumn{1}{l}{}     & 91.54 & 99.64                    & 90.17      & 14.08                        & 88.68     & 1.79                         & 89.19 & 1.21                     & 90.25            & 2.50             \\
                                  &  SBL w. Naive                                         & 92.58                & 100                      & 91.64 & 100                      & 91.64      & 100                          & 91.43     & 100                          & 91.43 & 100                      & 91.52            & 100             \\
                                  &  SBL w. EWC                                             & 92.58                & 100                      & 92.09 & 100                      & 91.90       & 100                          & 91.68     & 100                          & 91.55 & 100                      & 91.75            & 100             \\
                                 % &SBL w. Anchoring                                   & 92.58                & 100                      & 92.46 & 100                      & 91.88      & 100                          & 91.82     & 100                          & 91.67 & 100                      & 91.73            & 100             \\
                                  &  SBL w. AGEM                                            & 92.58                & 100                      & 92.22 & 100                      & 91.90       & 100                          & 91.67     & 100                          & 91.62 & 100                      & 91.74            & 100             \\
\hline
\multirow{5}{*}{\textbf{Blended}} & Original                                              & \multicolumn{1}{l}{} & \multicolumn{1}{l}{}     & 91.62 & 100                      & 91.07      & 58.30                         & 89.79     & 8.48                         & 89.50  & 22.29                    & 90.60             & 33.13           \\
                                  &  SBL w. Naive                                           & 91.91                & 100                      & 91.23 & 100                      & 91.25      & 100                          & 91.10     & 100                          & 91.19 & 100                      & 91.09            & 100             \\
                                  &  SBL w. EWC                                             & 91.91                & 100                      & 91.80 & 100                      & 91.78      & 100                          & 91.56     & 100                          & 91.44 & 99.98                    & 91.51            & 100             \\
                                 % &SBL w. Anchoring                                   & 91.91                & 100                      & 92.34 & 100                      & 91.75      & 99.98                        & 91.53     & 99.98                        & 91.45 & 100                      & 91.61            & 99.99           \\
                                  &  SBL w. AGEM                                            & 91.91                & 100                      & 91.86 & 100                      & 91.80      & 100                          & 91.54     & 100                          & 91.53 & 99.98                    & 91.55            & 100             \\
\hline
\multirow{5}{*}{\textbf{SIG}}     & Original                                              & \multicolumn{1}{l}{} & \multicolumn{1}{l}{}     & 91.22 & 99.94                    & 91.46      & 0.57                         & 89.59     & 0.38                         & 90.07 & 0.57                     & 90.22            & 0.51            \\
                                  &  SBL w. Naive                                           & 92.09                & 99.96                    & 91.29 & 99.06                    & 91.03      & 99.93                        & 91.06     & 95.98                        & 91.00 & 96.97                    & 90.96            & 97.94           \\
                                  &  SBL w. EWC                                             & 92.09                & 99.96                    & 91.81 & 99.38                    & 91.63      & 99.29                        & 91.69     & 97.96                        & 91.63 & 97.61                    & 91.73            & 99.28           \\
                                 % &SBL w. Anchoring                                   & 92.09                & 99.96                    & 92.26 & 99.80                    & 91.94      & 99.19                        & 91.76     & 96.19                        & 91.59 & 97.84                    & 91.53            & 98.74           \\
                                  &  SBL w. AGEM                                            & 92.09                & 99.96                    & 91.91 & 99.16                    & 91.61      & 98.19                        & 91.66     & 97.94                        & 91.59 & 97.17                    & 91.71            & 99.23          \\ \hline
\multirow{5}{*}{\textbf{Dynamic}}                      & Original                                              & \multicolumn{1}{l}{} & \multicolumn{1}{l}{}     & 91.15 & 99.96                    & 91.16      & 23.58                        & 89.91     & 6.93                         & 89.65      & 5.51       & 90.26            & 13.79           \\
                                                       & SBL w. Naive                                          & 91.96                & 99.94                    & 91.35 & 99.49                    & 91.63      & 100                          & 91.73     & 99.89                        & 91.64      & 99.85      & 91.22            & 99.89           \\
                                                       & SBL w. EWC                                            & 91.96                & 99.94                    & 91.49 & 99.99                    & 91.38      & 99.94                        & 91.24     & 99.94                        & 91.47      & 99.94      & 91.29            & 99.85           \\
                                                      % &SBL w. Anchoring                                      & 91.96                & 99.94                    & 91.98 & 100                      & 91.78      & 100                          & 91.78     & 99.99                        & 91.62      & 99.99      & 91.37            & 99.99           \\
                                                       & SBL w. AGEM                                           & 91.96                & 99.94                    & 91.81 & 100                      & 91.54      & 100                          & 91.69     & 99.89                        & 91.54      & 99.99      & 91.30            & 99.99          \\

\bottomrule
\end{tabular}
}
\end{table*}

\vspace{-0.5cm}

\begin{table}
\caption{The resilience against fine-tuning defenses in setting ResNet18 on GTSRB.}

\label{tab:main_gtsrb}
\resizebox{0.98\textwidth}{!}{
\small 
\centering
\begin{tabular}{c|l|cc|cc|cc|cc|cc|cc}
\toprule
\multirow{2}{*}{Attack} & \multicolumn{1}{c|}{\multirow{2}{*}{Training}} & \multicolumn{2}{c|}{Step 0}                     & \multicolumn{2}{c|}{Step 1}      & \multicolumn{2}{c|}{FT SGD-0.005} & \multicolumn{2}{c|}{FT SGD-0.01} & \multicolumn{2}{c|}{NAD}         & \multicolumn{2}{c}{FT-SAM} \\ \cline{3-14} 
                                  & \multicolumn{1}{c|}{}                                 & CA                   & \multicolumn{1}{c|}{ASR} & CA    & \multicolumn{1}{c|}{ASR} & CA         & \multicolumn{1}{c|}{ASR}     & CA        & \multicolumn{1}{c|}{ASR}     & CA    & \multicolumn{1}{c|}{ASR} & CA               & ASR             \\ \hline
\multirow{5}{*}{\textbf{Badnets}} & Original                                              & \multicolumn{1}{l}{} & \multicolumn{1}{l}{}     & 96.62 & 99.99                    & 97.30       & 93.31                        & 97.21     & 3.09                         & 97.17 & 1.03                     & 97.36            & 0.20             \\ 
                                  &   SBL w. Naive                                           & 98.15                & 100                      & 97.99 & 100                      & 97.65      & 100                          & 97.24     & 100                          & 96.59 & 100                      & 98.21            & 100             \\
                                  &   SBL w. EWC                                             & 98.15                & 100                      & 98.12 & 100                      & 98.04      & 100                          & 98.04     & 100                          & 97.95 & 100                      & 98.34            & 100             \\
                                 % &SBL w. Anchoring                                     & 98.15                & 100                      & 98.12 & 100                      & 98.07      & 100                          & 97.97     & 100                          & 98    & 100                      & 98.02            & 100             \\
                                  &   SBL w. AGEM                                            & 98.15                & 100                      & 98.16 & 100                      & 98.04      & 100                          & 98.04     & 100                          & 97.95 & 100                      & 98.25            & 100             \\
\hline
\multirow{5}{*}{\textbf{Blended}} & Original                                              & \multicolumn{1}{l}{} & \multicolumn{1}{l}{}     & 96.71 & 99.97                    & 97.19      & 52.74                        & 96.46     & 1.86                         & 96.15 & 0.15                     & 95.44            & 7.12            \\
                                  &   SBL w. Naive                                           & 98.32                & 100                      & 98.27 & 100                      & 98.13      & 100                          & 97.81     & 100                          & 97.84 & 99.84                    & 98.31            & 100             \\
                                  &   SBL w. EWC                                             & 98.32                & 100                      & 98.12 & 100                      & 98.21      & 100                          & 98.31     & 100                          & 98.21 & 100                      & 98.44            & 100             \\
                                 % &SBL w. Anchoring                                     & 98.32                & 100                      & 98.09 & 100                      & 98.16      & 100                          & 98.21     & 100                          & 98.20  & 100                      & 98.34            & 100             \\
                                  &   SBL w. AGEM                                            & 98.32                & 100                      & 98.21 & 100                      & 98.22      & 100                          & 98.31     & 100                          & 98.21 & 100                      & 98.42            & 100             \\
\hline
\multirow{5}{*}{\textbf{SIG}}     & Original                                              & \multicolumn{1}{l}{} & \multicolumn{1}{l}{}     & 96.47 & 99.99                    & 95.95      & 2.86                         & 93.41     & 0.14                         & 94.71 & 0                        & 95.47            & 1.29            \\
                                  &   SBL w. Naive                                           & 98.27                & 100                      & 98.23 & 99.99                    & 97.88      & 99.99                        & 97.67     & 99.92                        & 96.41 & 98.53                    & 98.12            & 100             \\
                                  &   SBL w. EWC                                             & 98.27                & 100                      & 98.14 & 100                      & 98.15      & 100                          & 98.13     & 100                          & 98.17 & 100                      & 98.11            & 100             \\
                                 % &SBL w. Anchoring                                     & 98.27                & 100                      & 98.16 & 100                      & 98.17      & 100                          & 98.13     & 100                          & 98.12 & 100                      & 98.00               & 100             \\
                                  &   SBL w. AGEM                                            & 98.27                & 100                      & 98.19 & 100                      & 98.15      & 100                          & 98.14     & 100                          & 98.17 & 100                      & 98.12            & 100            \\ \hline
\multirow{5}{*}{\textbf{Dynamic}}                      & Original                                              & \multicolumn{1}{l}{} & \multicolumn{1}{l}{}     & 96.17 & 99.98                    & 96.66      & 0.15                         & 95.23     & 0.02                         & 57.43      & 0.01       & 97.09            & 0.06            \\
                                                       & SBL w. Naive                                          & 98.37                & 100                      & 98.32 & 99.92                    & 98.06      & 100                          & 98.04     & 100                          & 98.05      & 100        & 98.19            & 100             \\
                                                       & SBL w. EWC                                            & 98.37                & 100                      & 98.23 & 100                      & 98.08      & 100                          & 90.02     & 100                          & 98.04      & 100        & 98.21            & 100             \\
                                                      % &SBL w. Anchoring                                      & 98.37                & 100                      & 98.23 & 100                      & 98.02      & 100                          & 97.99     & 100                          & 98.02      & 100        & 98.22            & 100             \\
                                                       & SBL w. AGEM                                           & 98.37                & 100                      & 98.22 & 100                      & 98.08      & 100                          & 98.03     & 100                          & 98.01      & 100        & 98.20            & 100   \\         

\bottomrule
\end{tabular}
}

\end{table}

% %%%%%%%%%%%%%%%%%%%%%%%%%%%%%%%%%%%%%%%%%%%%%
% %%%%%%%%%% RESNET-18 ON ImageNet-10 %%%%%%%%%%%%%%%
% %%%%%%%%%%%%%%%%%%%%%%%%%%%%%%%%%%%%%%%%%%%%%

% Please add the following required packages to your document preamble:
% \usepackage{multirow}
\begin{table}
\caption{The resilient against fine-tuning defenses in setting ResNet18 on ImageNet-10.}
\label{tab:main_imagenet10}
\resizebox{0.98\textwidth}{!}{
\small 
\centering
\begin{tabular}{c|l|cc|cc|cc|cc|cc|cc}
\toprule
\multirow{2}{*}{Attack} & \multicolumn{1}{c|}{\multirow{2}{*}{Training}} & \multicolumn{2}{c|}{Step 0}                     & \multicolumn{2}{c|}{Step 1}      & \multicolumn{2}{c|}{FT SGD-0.005} & \multicolumn{2}{c|}{FT SGD-0.01} & \multicolumn{2}{c|}{NAD}         & \multicolumn{2}{c}{FT-SAM} \\ \cline{3-14} 
                                  & \multicolumn{1}{c|}{}                                 & CA                   & \multicolumn{1}{c|}{ASR} & CA    & \multicolumn{1}{c|}{ASR} & CA         & \multicolumn{1}{c|}{ASR}     & CA        & \multicolumn{1}{c|}{ASR}     & CA    & \multicolumn{1}{c|}{ASR} & CA               & ASR             \\ \hline
\multirow{5}{*}{\textbf{Badnets}} & Original                                              & \multicolumn{1}{l}{} & \multicolumn{1}{l}{}     & 89.65 & 99.36                    & 73.27      & 6.71                         & 51.38     & 10.47                        & 41.03 & 5.23                     & 37.08            & 6.79            \\
                                  &   SBL w. Naive                                          & 89.46                & 99.91                    & 88.65 & 99.83                    & 85.69      & 91.54                        & 76.96     & 70.43                        & 69.77 & 73.68                    & 83.08            & 84.96           \\
                                  &   SBL w. EWC                                            & 89.46                & 99.91                    & 89.15 & 100                      & 87.08      & 99.83                        & 85.19     & 76.54                        & 73.69 & 74.03                    & 83.62            & 86.28           \\
                                  %% &SBL w. Anchoring                                      & 89.46                & 99.91                    & 89.88 & 100                      & 86.62      & 100                          & 84.04     & 87.95                        & 76.00 & 67.48                    & 82.00            & 76.62           \\
                                  &   SBL w. AGEM                                           & 89.46                & 99.91                    & 89.31 & 100                      & 87.27      & 99.96                        & 83.12     & 75.64                        & 70.23 & 71.07                    & 83.04            & 87.95           \\
\hline
\multirow{5}{*}{\textbf{Blended}} & Original                                              & \multicolumn{1}{l}{} & \multicolumn{1}{l}{}     & 89.12 & 99.70                    & 72.35      & 2.91                         & 59.00     & 6.88                         & 44.92 & 12.65                    & 66.08            & 1.41            \\
                                  &   SBL w. Naive                                          & 88.5                 & 98.8                     & 86.23 & 95.78                    & 84.85      & 79.10                        & 78.62     & 59.48                        & 72.16 & 64.53                    & 79.46            & 36.41           \\
                                  &   SBL w. EWC                                            & 88.5                 & 98.8                     & 88.12 & 97.35                    & 86.23      & 81.67                        & 81.88     & 74.06                        & 76.38 & 73.42                    & 82.96            & 46.79           \\
                                  %% &SBL w. Anchoring                                      & 88.5                 & 98.8                     & 89.23 & 97.78                    & 84.85      & 89.10                        & 84.27     & 62.48                        & 78.00 & 74.23                    & 82.96            & 46.79           \\
                                  &   SBL w. AGEM                                           & 88.5                 & 98.8                     & 88.31 & 97.74                    & 86.19      & 82.35                        & 83.85     & 69.02                        & 74.88 & 70.3                     & 81.73            & 62.95           \\
\hline
\multirow{5}{*}{\textbf{SIG}}     & Original                                              & \multicolumn{1}{l}{} & \multicolumn{1}{l}{}     & 89.27 & 99.87                    & 75.38      & 1.67                         & 55.50     & 7.82                         & 48.92 & 11.37                    & 63.31            & 5.00            \\
                                  &   SBL w. Naive                                          & 89.5                 & 99.83                    & 85.69 & 94.27                    & 85.58      & 69.15                        & 80.38     & 60.81                        & 66.31 & 45.21                    & 82.50            & 73.89           \\
                                  &   SBL w. EWC                                            & 89.5                 & 99.83                    & 89.23 & 99.83                    & 87.38      & 97.56                        & 83.00     & 76.79                        & 74.62 & 57.74                    & 81.38            & 88.29           \\
                                  %% &SBL w. Anchoring                                      & 89.5                 & 99.83                    & 89.31 & 99.70                    & 87.31      & 97.18                        & 83.62     & 64.53                        & 71.81 & 84.32                    & 82.15            & 66.54           \\
                                  &   SBL w. AGEM                                           & 89.5                 & 99.83                    & 89.23 & 99.83                    & 87.35      & 97.61                        & 82.92     & 76.84                        & 70.92 & 70.17                    & 80.81            & 87.35       
                         \\
\bottomrule
\end{tabular}
}

% \vspace{-0.3cm}
\end{table}

% \subsection{Loss Landscape Analysis}
% \label{sec:loss_analysis}

% In this section, we analyze the loss landscape properties of standard backdoor learning and our SBL framework using model interpolation. We conduct this analysis in two settings: ResNet18 on CIFAR-10 and ResNet18 on GTSRB, both with BadNets as the Attack. In the first column of Figure~\ref{fig:loss_analysis}, we visualize the loss and accuracy on the clean and poisoned test sets while linearly interpolating between backdoored and fine-tuned models in the CBL setup.
% While the clean loss and accuracy remain unchanged, the poisoning loss gradually increases, and the ASR decreases to nearly zero on the fine-tuned model. This indicates that with CBL, fine-tuning can effectively push the poisoned model out of the backdoor-affected area.
% On the other hand, the second and third columns of Figure~\ref{fig:loss_analysis} demonstrate that our SBL framework can identify a low-error pathway connecting backdoored models to the fine-tuned model. This finding is consistent with the backdoor loss analysis presented in section~\ref{sec:why_sbl_works}.

\begin{figure}[ht]
  \centering
  % Second row with two figures
  \begin{subfigure}{0.49\linewidth}
    \centering
    \includegraphics[width=\linewidth]{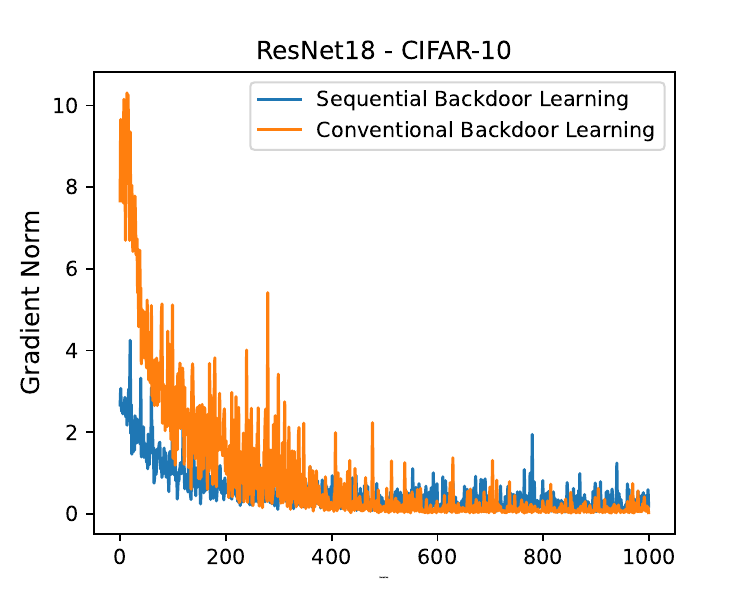}
    \label{fig:grad_norm_resnet18_cifar10}
  \end{subfigure}
  \hfill
  \begin{subfigure}{0.49\linewidth}
    \centering
    \includegraphics[width=\linewidth]{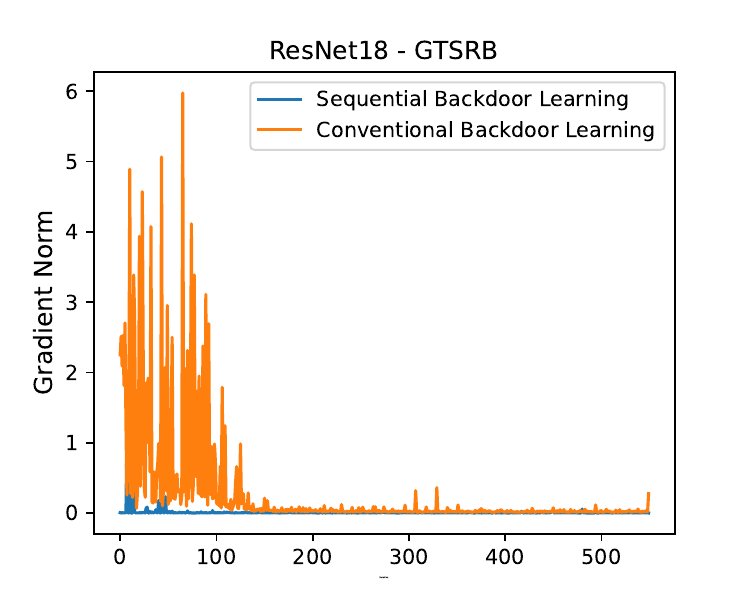}
    \label{fig:grad_norm_resnet18_gtsrb}
  \end{subfigure}

    \vspace{-0.5cm}
  \caption{Gradient norm comparison between conventional backdoor learning and our framework during defense fine-tuning with SGD-0.01 from the backdoored model.}
  \label{fig:grad_norm}
  \vspace{-0.5cm}
\end{figure}

% \subsection{Qualitative Analysis of SBL's Effectiveness}
% \label{sec:loss_analysis}
\vspace{0.2cm}
\noindent \textbf{Qualitative Analysis of SBL's Effectiveness.} To explain how the backdoors implanted by SBL can evade fine-tuning defenses, we examine the gradients' norm during the fine-tuning process for both CBL's and SBL's backdoored models. Figure~\ref{fig:grad_norm} shows the gradient norms of ResNet18 on CIFAR-10 and GTSRB. We can observe that in the early stage of fine-tuning, the gradient norm values of CBL are substantially higher than those of SBL. Higher values suggest that the fine-tuned model can more easily be pushed further away from the backdoored minimum, making it easier for the fine-tuning defenses to find backdoor-free local minima. 
% This suggests that it is easier for finetuning defenses to push the CBL's backdoored model into a different local minimum that is free of the backdoor. 

% This  which supports CBL poisoned model escape from backdoor regions. 

% Please add the following required packages to your document preamble:
% \usepackage{multirow}
\begin{table}
\caption{Ablation study on different architectures with CIFAR-10 dataset.}
\label{tab:arch_ablation_cifar10}
\resizebox{0.98\textwidth}{!}{
\centering

\small 

\begin{tabular}{c|l|C{1cm}C{1cm}|C{1cm}C{1cm}|C{1cm}C{1cm}|C{1cm}C{1cm}}
\toprule
\multirow{2}{*}{\textbf{CIFAR-10}}  & \multicolumn{1}{c|}{\multirow{2}{*}{Training}} & \multicolumn{2}{c|}{No Defense} & \multicolumn{2}{c|}{FT SGD-0.01} & \multicolumn{2}{c|}{NAD}         & \multicolumn{2}{c}{FT-SAM } \\
\cline{3-10}
                                    &                                                                & \multicolumn{1}{c}{CA}       & \multicolumn{1}{c|}{ASR}   & \multicolumn{1}{c}{CA}        & \multicolumn{1}{c|}{ASR}     & \multicolumn{1}{c}{CA}    & \multicolumn{1}{c|}{ASR} & \multicolumn{1}{c}{CA}               & ASR             \\ \hline
\multirow{6}{*}{\rotatebox[origin=c]{90}{\textbf{ResNet-20}}}  & Badnets                                              & 86.15    & 99.99                      &85.49     & 3.51                         & 85.07 & 2.21                     & 84.36            & 12.19           \\
& Badnets w. SBL                                             & 87.78    & 99.99                      & 86.96     & 99.44                        & 86.84 & 99.74                    & 86.74            & 98.93           \\ \cline{2-10} 
& Blended                                              & 87.08    & 99.9                       & 86.42     & 52.07                        & 86.19 & 25.52                    & 85.61            & 39.98           \\
& Blended w. SBL                                             & 87.77    & 99.98                      & 87.21     & 96.39                        & 86.83 & 96.12                    & 86.88            & 96.06           \\ \cline{2-10} 
& SIG                                              & 86.60    & 99.93                      & 85.92     & 1.18                         & 84.73 & 0.18                     & 84.73            & 1.21            \\
& SIG w. SBL                                             & 87.82    & 98.92                      & 86.98     & 88.77                        & 86.99 & 90.79                    & 87.33            & 88.89           \\
\midrule
\multirow{6}{*}{\rotatebox[origin=c]{90}{\textbf{VGG-16}}}    & Badnets                                              & 88.04    & 99.97                      & 86.93     & 3.20                          & 86.2  & 1.88                     & 88.12            & 1.34            \\
& Badnets w. SBL                                             & 90.19    & 100                        & 89.04     & 100                          & 89.18 & 100                      & 88.99            & 100             \\ \cline{2-10} 
& Blended                                              & 89.38    & 99.98                      & 88.15     & 9.73                         & 87.19 & 16.72                    & 88.33            & 11.74           \\
& Blended w. SBL                                             & 90.63    & 100                        & 90.12     & 100                          & 90.13 & 100                      & 90.25            & 100             \\ \cline{2-10} 
& SIG                                              & 88.26    & 98.88                      & 87.55     & 0.70                          & 86.29 & 0.41                     & 88.87            & 1.39            \\
& SIG w. SBL                                             & 90.16    & 99.99                      & 90.10     & 99.98                        & 90.12 & 99.99                    & 90.13            & 99.97 \\
\bottomrule
\end{tabular}
}

% \vspace{-0.5cm}
\end{table}

% \vspace{-0.5cm}
\subsection{Ablation Studies}
\label{sec:ablation_studies}

\noindent \textbf{Architecture Ablation.} We evaluate SBL's effectiveness on different network architectures, including VGG-16 and ResNet-20 (a lightweight version of ResNet). The experiments are conducted on CIFAR-10 and GTSRB with various attacks while SBL uses EWC. Tables~\ref{tab:arch_ablation_cifar10} and~\ref{tab:arch_ablation_gtsrb} show the backdoor's performance against various fine-tuning defenses. We observe similar effectiveness in SBL with these network architectures, confirming that SBL can successfully enhance the durability of the backdoored models against finetuning defenses.
% SBL is a general backdoor learning framework designed to enhance the resilience of poisoned models. 

% \vspace{-0.3cm}
% Please add the following required packages to your document preamble:
% \usepackage{multirow}
\begin{table}[ht]

\caption{Ablation study on different architectures with GTSRB dataset.}
\label{tab:arch_ablation_gtsrb}
\resizebox{0.98\textwidth}{!}{
\centering

\small 

\begin{tabular}{c|l|C{1cm}C{1cm}|C{1cm}C{1cm}|C{1cm}C{1cm}|C{1cm}C{1cm}}
\toprule
\multirow{2}{*}{\textbf{GTSRB}}  & \multicolumn{1}{c|}{\multirow{2}{*}{Training}} & \multicolumn{2}{c|}{No Defense} & \multicolumn{2}{c|}{FT SGD-0.01} & \multicolumn{2}{c|}{NAD}         & \multicolumn{2}{c}{FT-SAM } \\
\cline{3-10}
                                    &                                                                & \multicolumn{1}{c}{CA}       & \multicolumn{1}{c|}{ASR}   & \multicolumn{1}{c}{CA}        & \multicolumn{1}{c|}{ASR}     & \multicolumn{1}{c}{CA}    & \multicolumn{1}{c|}{ASR} & \multicolumn{1}{c}{CA}               & ASR             \\ \hline
\multirow{6}{*}{\rotatebox[origin=c]{90}{\textbf{ResNet-20}}} & Badnets                                              & 95.96    & 100                        & 95.87     & 1.36                         & 95.93 & 0                        & 94.96            & 0.06            \\
 & Badnets w. SBL                                             & 97.23    & 100                        & 97.06     & 100                          & 97.07 & 100                      & 97.21            & 100             \\ \cline{2-10} 
& Blended                                              & 95.70    & 99.89                      & 95.67     & 47.22                        & 95.69 & 73.54                    & 92.91            & 85.86           \\
& Blended w. SBL                                             & 97.14    & 100                        & 96.94     & 100                          & 96.97 & 100                      & 97.43            & 99.92           \\ \cline{2-10} 
& SIG                                              & 96.32    & 99.89                      & 94.76     & 6.23                         & 95.19 & 11.15                    & 93.52            & 40.52           \\
& SIG w. SBL                                             & 96.74    & 99.99                      & 96.92     & 99.98                        & 97.05 & 99.98                    & 96.98            & 99.98           \\
\midrule
\multirow{6}{*}{\rotatebox[origin=c]{90}{\textbf{VGG-16}}}    & Badnets                                              & 96.24    & 100                        & 95.50     & 3.80                          & 94.57 & 3.22                     & 95.96            & 4.61            \\
& Badnets w. SBL                                            & 97.58    & 100                        & 97.47     & 100                          & 97.52 & 100                      & 97.46            & 100             \\ \cline{2-10} 
                                    & Blended                                              & 96.25    & 100                        & 95.15     & 17.06                        & 93.37 & 6.48                     & 96.00               & 21.34           \\
& Blended w. SBL                                             & 97.17    & 100                        & 97.12     & 100                          & 97.25 & 100                      & 97.68            & 100             \\ \cline{2-10} 
& SIG                                              & 96.71    & 99.98                      & 94.69     & 0.62                         & 93.64 & 7.10                     & 95.46            & 64.18           \\
 & SIG w. SBL                                             & 96.90    & 100                        & 96.98     & 100                          & 96.96 & 100                      & 97.46            & 100             \\
\bottomrule
\end{tabular}
}

\end{table}
% \vspace{-0.3cm}

% \vspace{0.3cm}
\noindent \textbf{Role of Continual Learning.} As discussed in Section~\ref{sec:proposed_framework}, the CL design of SBL helps learn a backdoored model that is resistant to catastrophic forgetting of backdoor knowledge in a subsequent fine-tuning process.
To validate the importance of CL, we study the effectiveness of SBL with different CL techniques, including Naive, EWC, Anchoring, and AGEM, but \emph{without SAM} on CIFAR-10 and GTSRB. 
We present the results with BadNets attack in Table~\ref{tab:ab_cl}. As can be observed, leveraging CL can improve the resistance of a backdoored model against SGD-finetuning or NAD, compared to conventional training. Specifically, during learning the second task with benign samples, while naive fine-tuning (Naive) can increase the durability of the backdoor, fine-tuning with CL exhibits significantly better resistance.
On FT-SAM, however, the backdoors fail to persist. A possible explanation for the effectiveness of FT-SAM is that backdoor training with normal optimizers (e.g., SGD, Adam) creates backdoor-related neurons with high weights' norms~\cite{Zhu_2023_ICCV}, and FT-SAM additionally leverages SAM in its fine-tuning process to shrink the norms of these backdoor-related neurons. 
% This helps FT-SAM eliminate the backdoors more effectively. 

% \vspace{-0.5cm}
% Please add the following required packages to your document preamble:
% \usepackage{multirow}
\begin{table}

\caption{Ablation study on the role of Continual Learning with BadNets.}
\label{tab:ab_cl}
\resizebox{0.98\textwidth}{!}{
\small 
\centering
\begin{tabular}{c|l|C{1cm}C{1cm}|C{1cm}C{1cm}|C{1cm}C{1cm}|C{1cm}C{1cm}|C{1cm}C{1cm}|C{1cm}C{1cm}}
\toprule
\multicolumn{1}{c|}{\multirow{2}{*}{BadNets}} & \multicolumn{1}{c|}{Training} & \multicolumn{2}{c|}{Step 0} & \multicolumn{2}{c|}{Step 1} & \multicolumn{2}{c|}{FT-SGD 0.005} & \multicolumn{2}{c|}{FT-SGD 0.01} & \multicolumn{2}{c|}{NAD} & \multicolumn{2}{c}{FT-SAM} \\ \cline{3-14} 
\multicolumn{1}{c|}{} & \multicolumn{1}{c|}{w/o SAM} & CA & \multicolumn{1}{c|}{ASR} & CA & \multicolumn{1}{c|}{ASR} & CA & \multicolumn{1}{c|}{ASR} & CA & \multicolumn{1}{c|}{ASR} & CA & \multicolumn{1}{c|}{ASR} & CA & ASR \\ \hline
\multirow{5}{*}{\rotatebox[origin=c]{90}{\textbf{CIFAR-10}}} & Original &  &  & 91.54 & 99.64 & 90.17 & 14.08 & 88.68 & 1.79 & 89.19 & 1.21 & 90.25 & 2.5 \\
 & Naive & 90.2 & 100 & 91.66 & 99.99 & 91.46 & 93.71 & 91.21 & 23.19 & 90.88 & 42.22 & 91.51 & 2.84 \\
 & EWC & 90.2 & 100 & 91.69 & 99.99 & 91.36 & 96.31 & 91.01 & 40.52 & 90.84 & 80 & 91.27 & 6.13 \\
 & Anchoring & 90.2 & 100 & 91.65 & 99.99 & 91.46 & 99.58 & 91.15 & 85.73 & 91.06 & 58.39 & 91.22 & 9.34 \\
 & AGEM & 90.2 & 100 & 91.6 & 99.89 & 91.43 & 97.64 & 91.18 & 50.91 & 91.2 & 74.08 & 91.27 & 13.6 \\ \hline
\multirow{5}{*}{\rotatebox[origin=c]{90}{\textbf{GTSRB}}} & Original &  &  & 96.62 & 99.99 & 97.3 & 93.31 & 97.21 & 3.09 & 97.17 & 1.03 & 97.36 & 0.2 \\
 & Naive & 96.43 & 99.96 & 98.01 & 99.74 & 98.08 & 87.02 & 98.06 & 42.89 & 98.1 & 87.5 & 98.17 & 0.25 \\
 & EWC & 96.43 & 99.96 & 98.01 & 99.85 & 98.08 & 97.56 & 98.23 & 86.65 & 98.04 & 95.43 & 97.8 & 0.42 \\
 & Anchoring & 96.43 & 99.96 & 97.81 & 99.99 & 98.06 & 95.59 & 98.1 & 92.63 & 98.07 & 96.49 & 98.16 & 16.5 \\
 & AGEM & 96.43 & 99.96 & 98.03 & 98.76 & 98.08 & 99.38 & 98.15 & 94.25 & 98.12 & 98.53 & 97.88 & 1.61 \\ \bottomrule
\end{tabular}
}
\end{table}

% \vspace{-0.3cm}
% \begin{figure*}[htb]
%     \centering
%     \raisebox{0mm}{\rotatebox{90}{\tiny ResNet18 - CIFAR-10}}
%     \includegraphics[,clip,width=0.95\textwidth]{figures/badnet_cifar10_cl_ablation.pdf}%

%     \raisebox{0mm}{\rotatebox{90}{\tiny ResNet18 - GTSRB}}
%     \includegraphics[clip,width=0.95\textwidth]{figures/badnet_gtsrb_cl_ablation.pdf}
    
%     % \vspace{-3mm}
%     \caption{The Attack Success Rate (ASR) of backdoored models before and after defense with SGD - 0.01, NAD, and FT-SAM for setting ResNet18 on CIFAR-10 and GTSRB with BadNets poisoning.}
%     % \vspace{-5mm}
%     \label{fig:learning_ablation}
% \end{figure*}

% \vspace{-0.3cm}
\noindent \textbf{Role of Sharpness-Aware Minimizer.} In SBL, SAM is utilized to guide the backdoored model towards flat backdoor regions, which helps mitigate catastrophic forgetting of backdoors. We investigate the effectiveness of SAM training by utilizing it in both conventional backdoor training (CBL) and our framework.
The results are reported in Table~\ref{tab:ab_sam_on_cifar10}. As can be observed, the use of SAM can enhance the durability of the backdoored models, even when combined with standard backdoor training but only in some cases. The use of SAM in SBL, however, exhibits consistent backdoor durability against all fine-tuning defenses. 
% Notably, incorporating SAM into this traditional learning paradigm can effectively bypass the FT-SAM defense method~\cite{Zhu_2023_ICCV}. 
% We argue that this is because the poisoned model is trained to minimize the weight norms of neurons, directly countering the approach in FT-SAM.

\vspace{-0.5cm}
% Please add the following required packages to your document preamble:
% \usepackage{multirow}
\begin{table}

\caption{Ablation study on the role of SAM with ResNet-18 on CIFAR-10 setting.}
% \vspace{-0.2cm}
\label{tab:ab_sam_on_cifar10}
\resizebox{0.98\textwidth}{!}{
\small 
\centering
\begin{tabular}{c|l|C{1cm}C{1cm}|C{1cm}C{1cm}|C{1cm}C{1cm}|C{1cm}C{1cm}|C{1cm}C{1cm}}
\toprule
\multirow{2}{*}{Attack}   & \multicolumn{1}{c|}{\multirow{2}{*}{BadNets}} & \multicolumn{2}{c|}{No Defense}  & \multicolumn{2}{c|}{FT SGD 0.005} & \multicolumn{2}{c|}{FT SGD 0.01} & \multicolumn{2}{c|}{NAD}         & \multicolumn{2}{c}{FT-SAM} \\ \cline{3-12} 
                          & \multicolumn{1}{c|}{}                            & CA    & \multicolumn{1}{c|}{ASR} & CA     & \multicolumn{1}{c|}{ASR} & CA    & \multicolumn{1}{c|}{ASR} & CA    & \multicolumn{1}{c|}{ASR} & CA           & ASR         \\ \hline
\multirow{2}{*}{CBL} & w. SAM                                           & 90.20 & 100                      & 87.59  & 28.6                     & 88.22 & 3.14                     & 89.10 & 0.72                     & 90.49        & 99.99       \\
                          & w/o SAM                                          & 91.54 & 99.64                    & 90.17  & 14.08                    & 88.68 & 1.79                     & 89.19 & 1.21                     & 90.25        & 2.5         \\ \midrule
\multirow{2}{*}{SBL w. EWC}  & w. SAM                                           & 92.09 & 100                      & 91.90  & 100                      & 91.68 & 100                      & 91.55 & 100                      & 91.75        & 100         \\
                          & w/o SAM                                          & 91.69 & 99.99                    & 91.36  & 96.31                    & 91.01 & 40.52                    & 90.84 & 80                       & 91.27        & 6.13    \\ \bottomrule   
\end{tabular}
}
\end{table}

\vspace{-0.2cm}
\noindent \textbf{Ablation on learning the second task.}
To further understand the role of learning the second task, we investigate the impact of the learning rate on this process. We conduct experiments on CIFAR-10 with ResNet18, while  BadNets is the poisoning technique; we vary the value of the learning rate from 0.0005 to 0.01. In Table~\ref{tab:sec_lr}, we observe that, with smaller learning rates ($<$0.01) in Step 1, the backdoored model can effectively circumvent the fine-tuning based defenses. However, with larger learning rates, naively learning the second task will make it similar to the process of fine-tuning defense, resulting in a significant drop in ASR after the second task and the defense phase.
On the other hand, EWC remains effective in preventing the backdoor from being forgotten after learning the second task and the fine-tuning defense. However, this effectiveness comes at the cost of compromising clean performance. 
This experiment highlights the role of small learning rates and the effect of CL methods, in particular EWC, in mitigating backdoor forgetting.

% \vspace{-0.5cm}
% Please add the following required packages to your document preamble:
% \usepackage{multirow}
\begin{table}

\caption{The performance of backdoored model with different learning rates in the second task when training in Step 1 from $\theta_{B_0}$ to $\theta_B$ on CIFAR-10 with ResNet18.}
\label{tab:sec_lr}
\small 
\centering
\resizebox{0.85\textwidth}{!}{

\begin{tabular}{c|p{2.2cm}|C{1cm}C{1cm}|C{1cm}C{1cm}|C{1cm}C{1cm}}
\toprule
\multirow{2}{*}{Step 1 LR}       & \multicolumn{1}{c|}{\multirow{2}{*}{Training}} & \multicolumn{2}{c|}{Step 0}                   & \multicolumn{2}{c|}{Step 1} & \multicolumn{2}{c}{FT SGD-0.01} \\ \cline{3-8} 
                                 & \multicolumn{1}{c|}{}                                 & CA                     & ASR                  & CA           & ASR          & CA                 & ASR                \\ \hline
\multirow{2}{*}{\textbf{0.0005}} & SBL w. Naive                                                 & \multirow{8}{*}{92.22} & \multirow{8}{*}{100} & 91.69        & 100          & 91.66              & 100                \\
                                 & SBL w. EWC                                                   &                        &                      & 91.78        & 100          & 91.61              & 100                \\
\cline{1-2} \cline{5-8}
\multirow{2}{*}{\textbf{0.001}}  & SBL w. Naive                                                 &                        &                      & 91.64        & 100          & 91.52              & 100                \\
                                 & SBL w. EWC                                                   &                        &                      & 91.89        & 100          & 91.49              & 100                \\
\cline{1-2} \cline{5-8}
\multirow{2}{*}{\textbf{0.005}}  & SBL w. Naive                                                 &                        &                      & 91.23        & 100          & 90.57              & 100                \\
                                 & SBL w. EWC                                                   &                        &                      & 91.27        & 100          & 91.50              & 100                \\
\cline{1-2} \cline{5-8}
\multirow{2}{*}{\textbf{0.01}}   & SBL w. Naive                                                 &                        &                      & 89.76        & 7.07         & 89.76              & 6.34               \\
                                 & SBL w. EWC                                                   &                        &                      & 85.32        & 82.09        & 89.43              & 79.91             \\
\bottomrule
\end{tabular}
}
% \vspace{-0.3cm}
\vspace{-0.2cm}
\end{table}

\vspace{0.3cm}
\noindent \textbf{Ablation on poisoning rate.} Here, we perform ablation studies on the poisoning rate to understand the effectiveness of our SBL framework in generating resilient backdoored models. In the previous experiments, we poisoned 10\% of the training data; in the new experiments, we reduced this poisoning ratio to 5\% and 1\% while evaluating the resistance of the models on GTSRB and CIFAR-10. As we can observe in Table~\ref{tab:ab_pr}, even with 1\% poisoning ratio, training with SBL helps the backdoored model become resistant to fine-tuning defenses, including the state-of-the-art fine-tuning defense FT-SAM~\cite{Zhu_2023_ICCV}. 

\vspace{-0.5cm}
\begin{table}

\caption{Ablation on the poisoning rate (Pr) on our framework SBL and original training with BadNets }
\vspace{-0.1cm}
\label{tab:ab_pr}
\small
\centering
\begin{tabular}{c|c|l|C{0.8cm}C{0.8cm}|C{0.8cm}C{0.8cm}|C{0.8cm}C{0.8cm}|C{0.8cm}C{0.8cm}}
\toprule
\multirow{2}{*}{Dataset}  & \multirow{2}{*}{Pr}  & \multicolumn{1}{c|}{\multirow{2}{*}{Attack}} & \multicolumn{2}{c|}{No Defense} & \multicolumn{2}{c|}{FT SGD-0.01} & \multicolumn{2}{c|}{NAD}         & \multicolumn{2}{c}{FT-SAM } \\ \cline{4-11} 
                          &                      & \multicolumn{1}{c|}{}                        & CA             & ASR           & CA             & ASR            & CA         & ASR        & CA           & ASR         \\ \hline
\multirow{4}{*}{\rotatebox[origin=c]{90}{\textbf{CIFAR-10}}} & \multirow{2}{*}{1\%} & BadNets                                      & 90.82          & 75.11         & 90.18          & 2.98           & 89.89      & 2.16       & 90.63        & 4.01        \\
                          &                      & BadNets w. SBL                               & 92.18          & 100           & 91.69          & 99.90          & 91.70      & 99.90      & 92.04        & 99.80       \\ \cline{2-11}
                          & \multirow{2}{*}{5\%} & BadNets                                      & 90.27          & 100           & 89.16          & 1.74           & 89.20      & 1.73       & 90.42        & 3.38        \\
                          &                      & BadNets w. SBL                               & 91.95          & 100           & 91.80          & 100            & 91.65      & 100        & 91.94        & 100         \\ \midrule
\multirow{4}{*}{\rotatebox[origin=c]{90}{\textbf{GTSRB}}}    & \multirow{2}{*}{1\%} & BadNets                                      & 96.44          & 96.55         & 97.55          & 8.29           & 97.11      & 0.07       & 96.56        & 0.00        \\
                          &                      & BadNets w. SBL                               & 98.33          & 100           & 98.29          & 100            & 98.39      & 100        & 98.08        & 100         \\ \cline{2-11}
                          & \multirow{2}{*}{5\%} & BadNets                                      & 97.48          & 100           & 97.35          & 3.16           & 97.23      & 0.00       & 97.77        & 0.01        \\
                          &                      & BadNets w. SBL                               & 98.35          & 100           & 98.31          & 100            & 98.38      & 100        & 98.11        & 100        \\ \bottomrule
\end{tabular}
\end{table}

\vspace{-0.8cm}
\section{Conclusion}
In this paper, we approach backdoor attacks and defenses as continual learning tasks with a focus on mitigating backdoor forgetting. We introduce a novel backdoor training framework that can significantly intensify the resilience of the implanted backdoors when these models undergo different fine-tuning defenses. This framework splits the backdoor learning process into two steps with a sharpness-aware minimizer. This collaboration traps the poisoned model in the backdoor regions that are difficult for existing fine-tuning defenses to find alternative backdoor-free minima. We conduct extensive experiments on several benchmark backdoor datasets to demonstrate the effectiveness of our framework, compared to traditional backdoor learning. Our work exposes the existence of another significant backdoor threat against fine-tuning defenses, and we urge researchers to develop countermeasures for this type of attack.

% ---- Bibliography ----
%
% BibTeX users should specify bibliography style 'splncs04'.
% References will then be sorted and formatted in the correct style.
%
\bibliographystyle{splncs04}
\bibliography{main}

\clearpage
\setcounter{page}{1}
% \maketitlesupplementary

% \section{Rationale}
% \label{sec:rationale}
% % 
% Having the supplementary compiled together with the main paper means that:
% % 
% \begin{itemize}
% \item The supplementary can back-reference sections of the main paper, for example, we can refer to \cref{sec:intro};
% \item The main paper can forward reference sub-sections within the supplementary explicitly (e.g. referring to a particular experiment); 
% \item When submitted to arXiv, the supplementary will already included at the end of the paper.
% \end{itemize}
% % 
% To split the supplementary pages from the main paper, you can use \href{https://support.apple.com/en-ca/guide/preview/prvw11793/mac#:~:text=Delete%20a%20page%20from%20a,or%20choose%20Edit%20%3E%20Delete).}{Preview (on macOS)}, \href{https://www.adobe.com/acrobat/how-to/delete-pages-from-pdf.html#:~:text=Choose%20%E2%80%9CTools%E2%80%9D%20%3E%20%E2%80%9COrganize,or%20pages%20from%20the%20file.}{Adobe Acrobat} (on all OSs), as well as \href{https://superuser.com/questions/517986/is-it-possible-to-delete-some-pages-of-a-pdf-document}{command line tools}.

This document provides additional details, analysis and experimental results to support the main submission. We begin by providing additional details of the experimental setup in Section~\ref{sup:experimental_setup}. Then we provide additional results for the main experiment in Section~\ref{sec:sup_main_results}. 
Next, we test our SBL framework with pruning defense and compare it with the conventional backdoor learning on different architectures in Section~\ref{sec:sup_pruning}. 
Then, we provide more analysis on the model interpolation, gradient norm, and the number of defense epochs in Section~\ref{sec:sup_ml}, ~\ref{sec:sup_gn}, and~\ref{sec:sup_epochs}. 
Finally, we discuss the limitations in Section~\ref{sec:limitations}.

\section{Experimental Setup}
\label{sup:experimental_setup}

\subsection{Datasets}
\label{sup:training_dataset}

We use three benchmark datasets, namely CIFAR-10~\cite{krizhevsky2009learning}, GTSRB~\cite{Houben-IJCNN-2013}, and ImageNet-10 for our experiments. These datasets are widely used in various previous works, in both backdoor attacks and defenses.

\paragraph{CIFAR-10:} The dataset consists of 60,000 color images in 10 classes with the size 32 x 32. CIFAR-10 is divided into a training set and a test set with 50,000 and 10,000 images, respectively.
In training, we use random crop, and random horizontal flip as augmentation operators while no augmentation is applied during evaluation.

\paragraph{GTSRB:} The German Traffic Sign Recognition Benchmark - GTSRB consists of 60,000 images with 43 different classes and size varies from 32 x 32 to 250 x 250. The training set includes 39.209 images while the test set has 12,630 samples. We resize all images to 32 x 32 pixels. In training, we only augment samples with the random crop operator and do not use augmentation during testing.

\paragraph{ImageNet-10:} We follow \cite{huang2022learning} to select 10 classes from ImageNet-1K~\cite{imagenet15russakovsky}. There are 10,041 images in the training set and 2,600 images in the test set. We resize all images to 64 x 64 pixels. In training, we use random crop, random rotation, and random horizontal flip as augmentation operators while no augmentation is applied during evaluation.

\subsection{Training Details}
\label{sup:training_details}

We divide the training set into three sets: mixed set $\mathcal{D}_0$ (poisoned and benign samples), clean set $\mathcal{D}_1$, and defense set with portion 85\% - 10\% - 5\%, respectively.
In the testing, to generate the poisoning test set, we clone the clean test set, remove all clean samples with labels equal to the targeted class, and then poison all the images and change their labels to the targeted class.
We use the same classifier backbone ResNet18~\cite{he2016deep} for all datasets. We also do ablation studies on different architectures: VGG-16~\cite{simonyan2014very} and ResNet-20, a lightweight version of ResNet.
We use SGD optimizer for training the backdoored model and SAM~\cite{foret2021sharpnessaware} for training in Step 0. We choose $0.01$ learning rate for the first task and $0.001$ in Step 1. In Step 0, we train the backdoored model with 150 epochs and with 100 epochs in Step~1. 
In SAM, we set the value of hyperparameter $\rho$ is $0.05$ for both attack and defense.

\vspace{-0.5cm}
\paragraph{Poisoning Method:} 
We consider representative poisoning methods including BadNets~\cite{gu2017badnets}, Blended~\cite{chen2017targeted}, and SIG~\cite{barni2019new}. 
% In the case of BadNets, we apply a random color $3 \times 3$ square at the bottom-right corner as the backdoor trigger on all datasets. 
In all experiments, we poison 10\% of training data and set the targeted class to label 0.

\textbf{BadNets}~\cite{gu2017badnets} This is the first and most basic backdoor attack, in which the backdoor pattern is just a hand-picked image patch. We use a random color $3 \times 3$ square at the bottom-right corner as the backdoor trigger on all datasets.

\textbf{Blended}~\cite{chen2017targeted} This attack chooses an outer image, e.g., a Hello Kitty image, and blends it with clean samples to create backdoor data. The blended ratio is $20\%$.

\textbf{SIG}~\cite{barni2019new} This attack uses sinusoidal strips as a backdoor trigger. 

\textbf{Dynamic}~\cite{nguyen2020input} This attack uses a trigger generator to create different triggers for different inputs.

\paragraph{Defense Methods:} 
We evaluate the persistence of backdoored models against fine-tuning-based defense methods, including standard fine-tuning~\cite{liu2018fine}, SAM-FT~\cite{Zhu_2023_ICCV} and NAD~\cite{li2021neural}. 
% We use SGD to fine-tune with two learning rates $0.01$ and $0.005$,  SAM-FT~\cite{Zhu_2023_ICCV} with learning rate $0.005$. We fine-tune the model with 50 epochs. With NAD, we fine-tune teacher and student with $0.01$ learning rate on 20 epochs. 

\textbf{Standard Fine-tuning}: Fine-tuning is a transfer learning technique that allows a pre-trained model to adapt to new tasks using new data. Fine-tuning a poisoned model on clean data can help mitigate backdoor attacks due to catastrophic forgetting. In our experiments, we adopt whole-model fine-tuning on unseen clean data on 50 epochs. We select two different learning-rate values, 0.01 and 0.005, to evaluate the influence of the learning rate choice on defense performance. 

\textbf{Neural Attentive Distillation (NAD):}~\cite{li2021neural} argue that fine-tuning alone is inadequate to defend against backdoor attacks, NAD uses knowledge distillation~\cite{hinton2015distilling} technique for backdoor mitigation. It first fine-tunes the backdoored model to obtain a teacher model and then uses the teacher to guide the fine-tuning of the backdoored student model.
We set $0.01$ as the learning rate and train both teacher and student models with 20 epochs.

\textbf{SAM-FT}~\cite{Zhu_2023_ICCV} observes that backdoor-related neurons often have larger norms, Zhu et. al., \cite{Zhu_2023_ICCV} proposed to incorporate fine-tuning with Sharpness-Aware Minimizer (SAM)~\cite{foret2021sharpnessaware} to shrink the norms of these neurons.
We fine-tune on 50 epochs with a learning rate of 0.005.

%%%%%%%%%%%%%%%%%%%%%%%%%%%%%%%%%%%%%%%%%%%%%
%%%%%%%%%% RESNET-18 ON CIFAR-10 %%%%%%%%%%%%%%%
%%%%%%%%%%%%%%%%%%%%%%%%%%%%%%%%%%%%%%%%%%%%%

% Please add the following required packages to your document preamble:
% \usepackage{multirow}
\begin{table*}[]
\resizebox{0.98\textwidth}{!}{
\small 
\centering
\begin{tabular}{c|l|cc|cc|cc|cc|cc|cc}
\toprule
\multirow{2}{*}{Poisoning method} & \multicolumn{1}{c|}{\multirow{2}{*}{Training method}} & \multicolumn{2}{c|}{Step 0}                     & \multicolumn{2}{c|}{Step 1}      & \multicolumn{2}{c|}{FT w. SGD-0.005} & \multicolumn{2}{c|}{FT w. SGD-0.01} & \multicolumn{2}{c|}{NAD}         & \multicolumn{2}{c}{FT-SAM} \\ \cline{3-14} 
                                  & \multicolumn{1}{c|}{}                                 & CA                   & \multicolumn{1}{c|}{ASR} & CA    & \multicolumn{1}{c|}{ASR} & CA         & \multicolumn{1}{c|}{ASR}     & CA        & \multicolumn{1}{c|}{ASR}     & CA    & \multicolumn{1}{c|}{ASR} & CA               & ASR             \\ \hline
\multirow{5}{*}{\textbf{Badnets}} & Original                                              & \multicolumn{1}{l}{} & \multicolumn{1}{l}{}     & 91.54 & 99.64                    & 90.17      & 14.08                        & 88.68     & 1.79                         & 89.19 & 1.21                     & 90.25            & 2.50             \\
                                  &  SBL w. Naive                                         & 92.58                & 100                      & 91.64 & 100                      & 91.64      & 100                          & 91.43     & 100                          & 91.43 & 100                      & 91.52            & 100             \\
                                  &  SBL w. EWC                                             & 92.58                & 100                      & 92.09 & 100                      & 91.90       & 100                          & 91.68     & 100                          & 91.55 & 100                      & 91.75            & 100             \\
                                  &  SBL w. Anchoring                                   & 92.58                & 100                      & 92.46 & 100                      & 91.88      & 100                          & 91.82     & 100                          & 91.67 & 100                      & 91.73            & 100             \\
                                  &  SBL w. AGEM                                            & 92.58                & 100                      & 92.22 & 100                      & 91.90       & 100                          & 91.67     & 100                          & 91.62 & 100                      & 91.74            & 100             \\
\hline
\multirow{5}{*}{\textbf{Blended}} & Original                                              & \multicolumn{1}{l}{} & \multicolumn{1}{l}{}     & 91.62 & 100                      & 91.07      & 58.30                         & 89.79     & 8.48                         & 89.50  & 22.29                    & 90.60             & 33.13           \\
                                  &  SBL w. Naive                                           & 91.91                & 100                      & 91.23 & 100                      & 91.25      & 100                          & 91.10     & 100                          & 91.19 & 100                      & 91.09            & 100             \\
                                  &  SBL w. EWC                                             & 91.91                & 100                      & 91.80 & 100                      & 91.78      & 100                          & 91.56     & 100                          & 91.44 & 99.98                    & 91.51            & 100             \\
                                  &  SBL w. Anchoring                                   & 91.91                & 100                      & 92.34 & 100                      & 91.75      & 99.98                        & 91.53     & 99.98                        & 91.45 & 100                      & 91.61            & 99.99           \\
                                  &  SBL w. AGEM                                            & 91.91                & 100                      & 91.86 & 100                      & 91.80      & 100                          & 91.54     & 100                          & 91.53 & 99.98                    & 91.55            & 100             \\
\hline
\multirow{5}{*}{\textbf{SIG}}     & Original                                              & \multicolumn{1}{l}{} & \multicolumn{1}{l}{}     & 91.22 & 99.94                    & 91.46      & 0.57                         & 89.59     & 0.38                         & 90.07 & 0.57                     & 90.22            & 0.51            \\
                                  &  SBL w. Naive                                           & 92.09                & 99.96                    & 91.29 & 99.06                    & 91.03      & 99.93                        & 91.06     & 95.98                        & 91.00 & 96.97                    & 90.96            & 97.94           \\
                                  &  SBL w. EWC                                             & 92.09                & 99.96                    & 91.81 & 99.38                    & 91.63      & 99.29                        & 91.69     & 97.96                        & 91.63 & 97.61                    & 91.73            & 99.28           \\
                                  &  SBL w. Anchoring                                   & 92.09                & 99.96                    & 92.26 & 99.80                    & 91.94      & 99.19                        & 91.76     & 96.19                        & 91.59 & 97.84                    & 91.53            & 98.74           \\
                                  &  SBL w. AGEM                                            & 92.09                & 99.96                    & 91.91 & 99.16                    & 91.61      & 98.19                        & 91.66     & 97.94                        & 91.59 & 97.17                    & 91.71            & 99.23          \\
\bottomrule
\end{tabular}
}

\caption{The resilience against fine-tuning defenses in setting  ResNet18 on CIFAR-10. Results with our SBL framework are in the shade.}
\label{tab:sup_cifar10}
\end{table*}

%%%%%%%%%%%%%%%%%%%%%%%%%%%%%%%%%%%%%%%%%%%%%
%%%%%%%%%% RESNET-18 ON GTSRB %%%%%%%%%%%%%%%
%%%%%%%%%%%%%%%%%%%%%%%%%%%%%%%%%%%%%%%%%%%%%

% Please add the following required packages to your document preamble:
% \usepackage{multirow}
\begin{table*}[]
\resizebox{0.98\textwidth}{!}{
\small 
\centering
\begin{tabular}{c|l|cc|cc|cc|cc|cc|cc}
\toprule
\multirow{2}{*}{Poisoning method} & \multicolumn{1}{c|}{\multirow{2}{*}{Training method}} & \multicolumn{2}{c|}{Step 0}                     & \multicolumn{2}{c|}{Step 1}      & \multicolumn{2}{c|}{FT w. SGD-0.005} & \multicolumn{2}{c|}{FT w. SGD-0.01} & \multicolumn{2}{c|}{NAD}         & \multicolumn{2}{c}{FT-SAM} \\ \cline{3-14} 
                                  & \multicolumn{1}{c|}{}                                 & CA                   & \multicolumn{1}{c|}{ASR} & CA    & \multicolumn{1}{c|}{ASR} & CA         & \multicolumn{1}{c|}{ASR}     & CA        & \multicolumn{1}{c|}{ASR}     & CA    & \multicolumn{1}{c|}{ASR} & CA               & ASR             \\ \hline
\multirow{5}{*}{\textbf{Badnets}} & Original                                              & \multicolumn{1}{l}{} & \multicolumn{1}{l}{}     & 96.62 & 99.99                    & 97.30       & 93.31                        & 97.21     & 3.09                         & 97.17 & 1.03                     & 97.36            & 0.20             \\ 
                                  &   SBL w. Naive                                           & 98.15                & 100                      & 97.99 & 100                      & 97.65      & 100                          & 97.24     & 100                          & 96.59 & 100                      & 98.21            & 100             \\
                                  &   SBL w. EWC                                             & 98.15                & 100                      & 98.12 & 100                      & 98.04      & 100                          & 98.04     & 100                          & 97.95 & 100                      & 98.34            & 100             \\
                                  &   SBL w. Anchoring                                     & 98.15                & 100                      & 98.12 & 100                      & 98.07      & 100                          & 97.97     & 100                          & 98    & 100                      & 98.02            & 100             \\
                                  &   SBL w. AGEM                                            & 98.15                & 100                      & 98.16 & 100                      & 98.04      & 100                          & 98.04     & 100                          & 97.95 & 100                      & 98.25            & 100             \\
\hline
\multirow{5}{*}{\textbf{Blended}} & Original                                              & \multicolumn{1}{l}{} & \multicolumn{1}{l}{}     & 96.71 & 99.97                    & 97.19      & 52.74                        & 96.46     & 1.86                         & 96.15 & 0.15                     & 95.44            & 7.12            \\
                                  &   SBL w. Naive                                           & 98.32                & 100                      & 98.27 & 100                      & 98.13      & 100                          & 97.81     & 100                          & 97.84 & 99.84                    & 98.31            & 100             \\
                                  &   SBL w. EWC                                             & 98.32                & 100                      & 98.12 & 100                      & 98.21      & 100                          & 98.31     & 100                          & 98.21 & 100                      & 98.44            & 100             \\
                                  &   SBL w. Anchoring                                     & 98.32                & 100                      & 98.09 & 100                      & 98.16      & 100                          & 98.21     & 100                          & 98.20  & 100                      & 98.34            & 100             \\
                                  &   SBL w. AGEM                                            & 98.32                & 100                      & 98.21 & 100                      & 98.22      & 100                          & 98.31     & 100                          & 98.21 & 100                      & 98.42            & 100             \\
\hline
\multirow{5}{*}{\textbf{SIG}}     & Original                                              & \multicolumn{1}{l}{} & \multicolumn{1}{l}{}     & 96.47 & 99.99                    & 95.95      & 2.86                         & 93.41     & 0.14                         & 94.71 & 0                        & 95.47            & 1.29            \\
                                  &   SBL w. Naive                                           & 98.27                & 100                      & 98.23 & 99.99                    & 97.88      & 99.99                        & 97.67     & 99.92                        & 96.41 & 98.53                    & 98.12            & 100             \\
                                  &   SBL w. EWC                                             & 98.27                & 100                      & 98.14 & 100                      & 98.15      & 100                          & 98.13     & 100                          & 98.17 & 100                      & 98.11            & 100             \\
                                  &   SBL w. Anchoring                                     & 98.27                & 100                      & 98.16 & 100                      & 98.17      & 100                          & 98.13     & 100                          & 98.12 & 100                      & 98.00               & 100             \\
                                  &   SBL w. AGEM                                            & 98.27                & 100                      & 98.19 & 100                      & 98.15      & 100                          & 98.14     & 100                          & 98.17 & 100                      & 98.12            & 100            \\
\bottomrule
\end{tabular}
}

\caption{The resilience against fine-tuning defenses in setting ResNet18 on GTSRB. Results with our SBL framework are in the shade.}
\vspace{-0.5cm}
\label{tab:sup_gtsrb}
\end{table*}

%%%%%%%%%%%%%%%%%%%%%%%%%%%%%%%%%%%%%%%%%%%%%
%%%%%%%%%% RESNET-18 ON ImageNet-10 %%%%%%%%%%%%%%%
%%%%%%%%%%%%%%%%%%%%%%%%%%%%%%%%%%%%%%%%%%%%%

% Please add the following required packages to your document preamble:
% \usepackage{multirow}
\begin{table*}[]
\resizebox{0.98\textwidth}{!}{
\small 
\centering
\begin{tabular}{c|l|cc|cc|cc|cc|cc|cc}
\toprule
\multirow{2}{*}{Poisoning method} & \multicolumn{1}{c|}{\multirow{2}{*}{Training method}} & \multicolumn{2}{c|}{Step 0}                     & \multicolumn{2}{c|}{Step 1}      & \multicolumn{2}{c|}{FT w. SGD-0.005} & \multicolumn{2}{c|}{FT w. SGD-0.01} & \multicolumn{2}{c|}{NAD}         & \multicolumn{2}{c}{SAM Finetuning} \\ \cline{3-14} 
                                  & \multicolumn{1}{c|}{}                                 & CA                   & \multicolumn{1}{c|}{ASR} & CA    & \multicolumn{1}{c|}{ASR} & CA         & \multicolumn{1}{c|}{ASR}     & CA        & \multicolumn{1}{c|}{ASR}     & CA    & \multicolumn{1}{c|}{ASR} & CA               & ASR             \\ \hline
\multirow{5}{*}{\textbf{Badnets}} & Original                                              & \multicolumn{1}{l}{} & \multicolumn{1}{l}{}     & 89.65 & 99.36                    & 73.27      & 6.71                         & 51.38     & 10.47                        & 41.03 & 5.23                     & 37.08            & 6.79            \\
                                  &   SBL w. Naive                                          & 89.46                & 99.91                    & 88.65 & 99.83                    & 85.69      & 91.54                        & 76.96     & 70.43                        & 69.77 & 73.68                    & 83.08            & 84.96           \\
                                  &   SBL w. EWC                                            & 89.46                & 99.91                    & 89.15 & 100                      & 87.08      & 99.83                        & 85.19     & 76.54                        & 73.69 & 74.03                    & 83.62            & 86.28           \\
                                  &   SBL w. Anchoring                                      & 89.46                & 99.91                    & 89.88 & 100                      & 86.62      & 100                          & 84.04     & 87.95                        & 76.00 & 67.48                    & 82.00            & 76.62           \\
                                  &   SBL w. AGEM                                           & 89.46                & 99.91                    & 89.31 & 100                      & 87.27      & 99.96                        & 83.12     & 75.64                        & 70.23 & 71.07                    & 83.04            & 87.95           \\
\hline
\multirow{5}{*}{\textbf{Blended}} & Original                                              & \multicolumn{1}{l}{} & \multicolumn{1}{l}{}     & 89.12 & 99.70                    & 72.35      & 2.91                         & 59.00     & 6.88                         & 44.92 & 12.65                    & 66.08            & 1.41            \\
                                  &   SBL w. Naive                                          & 88.5                 & 98.8                     & 86.23 & 95.78                    & 84.85      & 79.10                        & 78.62     & 59.48                        & 72.16 & 64.53                    & 79.46            & 36.41           \\
                                  &   SBL w. EWC                                            & 88.5                 & 98.8                     & 88.12 & 97.35                    & 86.23      & 81.67                        & 81.88     & 74.06                        & 76.38 & 73.42                    & 82.96            & 46.79           \\
                                  &   SBL w. Anchoring                                      & 88.5                 & 98.8                     & 89.23 & 97.78                    & 84.85      & 89.10                        & 84.27     & 62.48                        & 78.00 & 74.23                    & 82.96            & 46.79           \\
                                  &   SBL w. AGEM                                           & 88.5                 & 98.8                     & 88.31 & 97.74                    & 86.19      & 82.35                        & 83.85     & 69.02                        & 74.88 & 70.3                     & 81.73            & 62.95           \\
\hline
\multirow{5}{*}{\textbf{SIG}}     & Original                                              & \multicolumn{1}{l}{} & \multicolumn{1}{l}{}     & 89.27 & 99.87                    & 75.38      & 1.67                         & 55.50     & 7.82                         & 48.92 & 11.37                    & 63.31            & 5.00            \\
                                  &   SBL w. Naive                                          & 89.50                 & 99.83                    & 85.69 & 94.27                    & 85.58      & 69.15                        & 80.38     & 60.81                        & 66.31 & 45.21                    & 82.50            & 73.89           \\
                                  &   SBL w. EWC                                            & 89.5                 & 99.83                    & 89.23 & 99.83                    & 87.38      & 97.56                        & 83.00     & 76.79                        & 74.62 & 57.74                    & 81.38            & 88.29           \\
                                  &   SBL w. Anchoring                                      & 89.5                 & 99.83                    & 89.31 & 99.70                    & 87.31      & 97.18                        & 83.62     & 64.53                        & 71.81 & 84.32                    & 82.15            & 86.54           \\
                                  &   SBL w. AGEM                                           & 89.5                 & 99.83                    & 89.23 & 99.83                    & 87.35      & 97.61                        & 82.92     & 76.84                        & 70.92 & 70.17                    & 80.81            & 87.35       
                         \\
\bottomrule
\end{tabular}
}

\caption{The resilience against fine-tuning defenses in setting  ResNet18 on ImageNet-10. Results with our SBL framework are in the shade.}
\label{tab:sup_imagenet10}
\vspace{-0.3cm}
\end{table*}

\subsection{Detailed Baselines}

\vspace{-0.2cm}
\paragraph{Baselines}
Here, along with original backdoor training, we select several CL techniques to incorporate in our framework to train backdoored model including Naive, EWC~\cite{kirkpatrick2017overcoming}, Anchoring~\cite{zhang2022how}, and AGEM~\cite{chaudhry2018efficient}.

\textbf{Original}: This is a conventional backdoor training. Particularly, we incorporate mixed dataset $\mathcal{D}_0$ and clean data $\mathcal{D}_1$ into one to train the backdoored model. 

\textbf{Naive}: Learning new task without CL techniques.

\textbf{EWC}~\cite{kirkpatrick2017overcoming}: This is a regularization CL method that identifies important weights via the Fisher Information Matrix (FIM). 
EWC penalizes changes in crucial weights, the regularization term is $\sum_j F_j (\theta_j - \theta_{B_0 j})^2$ where $F_j$ is the $j$-th element in the diagonal of FIM which is an approximation of $H_{jj}$ in the Hessian matrix $H$ calculated on dataset $D_0$.

\textbf{Anchoring}~\cite{zhang2022how}: This method is used in~\cite{zhang2022how} to force the model to output similar logits on clean data. It adds regularization term: $\sum_i^C (s_i(x; \theta) - s_i(x; \theta_{B_0}))$ where $C$ is set of classes, $x$ is benign sample, and $s_i(.)$ is the logit for class $i$.

\textbf{AGEM}~\cite{chaudhry2018efficient}: This is a memory-based CL method that stores a buffer of data from the first task. In updating the model on the second task, the gradient update is projected in a direction that does not hamper the update of previous tasks. In particular, let $g$ be the gradient computed with the incoming mini-batch and $g_{ref}$ be the gradient computed with the same size mini-batch randomly selected from the memory buffer. 
In A-GEM, if $g^{\top}g_{ref} \geq 0$, $g$  is used for gradient update but when $g^{\top}g_{ref} < 0$, $g$ is projected such that $g^{\top} g_{ref} = 0$. The gradient after projection is: $\tilde{g} = g - \frac{g^{\top} g_{ref}}{g_{ref}^{\top} g_{ref}} g_{ref}$. In our experiments, we store 512 samples in memory which is taken from the poisoned set $D_0$. In each iteration when training on Step 1, we randomly select 128 samples from memory which is used to calculate $g_{ref}$.

%%%%%%%%%%%%%
%%%%%%%
%%
% \subsection{Why should our method work in practice?}
\subsection{Further analysis of SBL} \label{appx-sec:why_sbl_works}

%While our method SBL looks simple and is easy to implement, experiments (Section \ref{sec:experiments}) show that it is surprisingly effective in training backdoor models resilient  against powerful fine-tuning defense methods. 
% To better understand the effectiveness of SBL, we discuss here motivations and heuristic explanations for its working mechanism.
%In particular, we seek to answer two questions: $(i)$ What properties that models are expected to attain under SBL training? and $(ii)$ Why should SBL be able to produce such models?

%SBL's objective is to cause the model to converge to flat backdoored regions that makes the attacks resilient to fine-tuning defenses; i.e., the finetuned model is still trapped in the flat region of backdoor knowledge. In this section, we provide an analysis to support this intuition. 

%We denote the two steps of SBL as multi-task (MT) training followed by continual learning (CL).

Recall that SBL first trains the backdoored model on both clean and poisoned data to obtain $\theta_{B_0}$ (MT), then fine-tune this model from $\theta_{B_0}$ with clean data and a tiny learning rate to achieve $\theta_B$ (CL). Denote by$\theta_{F}$ a fine-tuned model from $\theta_B$. 

Our experiments show that the training method in \textbf{Step 1} of Algorithm~\ref{alg:sbl_framework} of SBL is able to find low-loss minima for clean data, while still maintaining the performance on poisoned data. 
This can be heuristically explained using Taylor approximation of the backdoor loss $\mathcal{L}_0(\theta_B):= \mathcal{L}(D_0; \theta_B)$ as follows.
%Following the way in~\cite{mirzadeh2020understanding}, it can be formulated as:
Since $\theta_{B_0}$ is the optimal weight, we can assume that the gradient vanishes $\nabla \mathcal{L}_0 (\theta_{B_0}) \approx 0$ in the second order approximation:
{\scriptsize
\begin{align*}
    \mathcal{L}_0(\theta_B) \approx \mathcal{L}_0(\theta_{B_0}) + (\theta_B - \theta_{B_0})^{\top} \nabla \mathcal{L}_0 (\theta_{B_0}) 
    + \frac{1}{2} (\theta_B - \theta_{B_0})^{\top} \nabla^2 \mathcal{L}_0 (\theta_{B_0}) (\theta_B - \theta_{B_0}) 
    % \le \mathcal{L}_0(\theta_{B_0}) + \frac{1}{2} \lambda_0^{max} ||\theta_B - \theta_{B_0}||^2
\end{align*}
}
from which we obtain the estimate:
\begin{equation}
    \small
    \mathcal{L}_0(\theta_B) - \mathcal{L}_0(\theta_{B_0}) \le \frac{1}{2} \lambda_{B_0}^{max} ||\theta_B - \theta_{B_0}||^2,
    \label{eq:loss_bound_1}
\end{equation}
where  $\lambda_{B_0}^{max}$ is the maximum eigenvalue of $\nabla^2 \mathcal{L}_0 (\theta_{B_0})$. 
Since $(i)$ the change of weight is low ($||\theta_B - \theta_{B_0}|| \approx 0$) due to the \textit{tiny learning rate} when continuously fine-tuning on clean data in \textbf{Step 1}, and 
$(ii)$ $| \lambda_{B_0}^{max}| \approx 0$ due to the effect of the flatness-aware optimizer SAM~\cite{foret2021sharpnessaware} that we use in Step 0, Eq.~(\ref{eq:loss_bound_1}) implies that the change in backdoor loss after Step 1 of SBL is small, as we empirically observed in Fig.~\ref{fig:loss_analysis}.

We have seen in Fig.~\ref{fig:loss_analysis} that the backdoored model $\theta_B$ trained with our method SBL already converges in low-loss for clean data, which makes it hard to escape from the current region of weights when applying fine-tuning defenses.
Again, we can use Taylor approximation on the backdoor loss \textit{after fine-tuning} defenses (from $\theta_B$ to $\theta_F$) in a similar fashion
%\begin{align*}
%    \mathcal{L}_0(\theta_F) \approx \mathcal{L}_0(\theta_{B}) + (\theta_F - \theta_B)^{\top} \nabla \mathcal{L}_0 (\theta_B) \\
%    + \frac{1}{2} (\theta_F - \theta_B)^{\top} \nabla^2 \mathcal{L}_0 (\theta_B) (\theta_F - \theta_B)
    % \le \mathcal{L}_0(\theta_{B_0}) + \frac{1}{2} \lambda_1^{max} ||\theta_B - \theta_{B_0}||^2
%\end{align*} 
to obtain:
\begin{equation}
\small
    \mathcal{L}_0(\theta_F) - \mathcal{L}_0(\theta_B) \le \frac{1}{2} \lambda_{B}^{max} ||\theta_F - \theta_B||^2.
    \label{eq:loss_bound_2}
\end{equation}

Since the backdoored model $\theta_B$ is optimized on clean data in \textbf{Step 1} of SBL, the gradient norm will be very small when fine-tuning on clean data afterwards. 
At the same time, the flatness property keeps $|\lambda_{B}^{max}| \approx 0$.
Consequently, Eq.~\ref{eq:loss_bound_2} implies that there is only a tiny change in the model's parameters during fine-tuning defenses. 
This is inline with our empirical observation in Fig.~\ref{fig:grad_norm}. 
As a result, the fine-tuned models are still trapped in the backdoor knowledge region.

%%
%%%%%%%%
%%%%%%%%%%%%%

\section{Supplementary Experimental Results}

\subsection{Additional Main Results}
\label{sec:sup_main_results}

Here, we provide additional results when incorporating Anchoring with our framework in three settings. We show results in Table~\ref{tab:sup_cifar10},~\ref{tab:sup_gtsrb}, and~\ref{tab:sup_imagenet10}. 

\subsection{Additional Results with Pruning Defense}
\label{sec:sup_pruning}
\begin{figure}
  \centering
  % Second row with two figures
  % \raisebox{12mm}{\rotatebox{90}{\tiny ResNet18}}%
  \begin{subfigure}{0.48\linewidth}
    \centering
    \includegraphics[width=\linewidth]{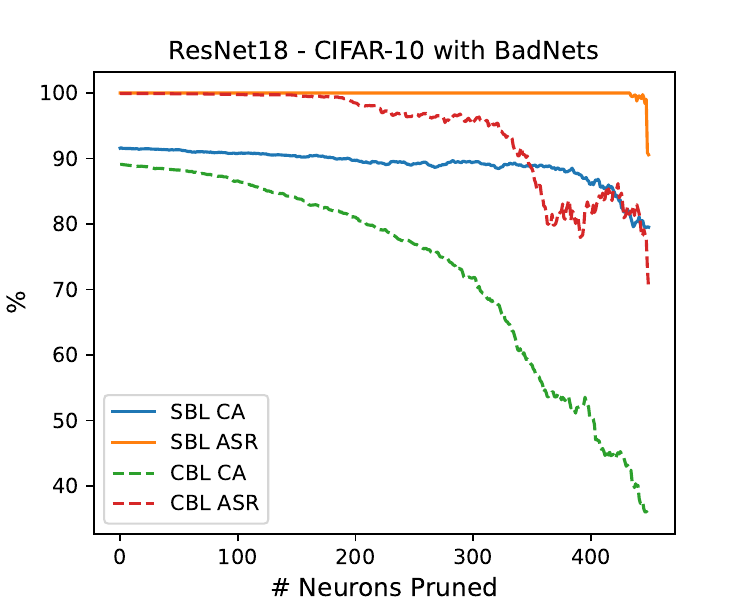}
    % \label{fig:MI_cifar_badnet_joint}
  \end{subfigure}
    % \hfill
    \begin{subfigure}{0.48\linewidth}
    \centering
    \includegraphics[width=\linewidth]{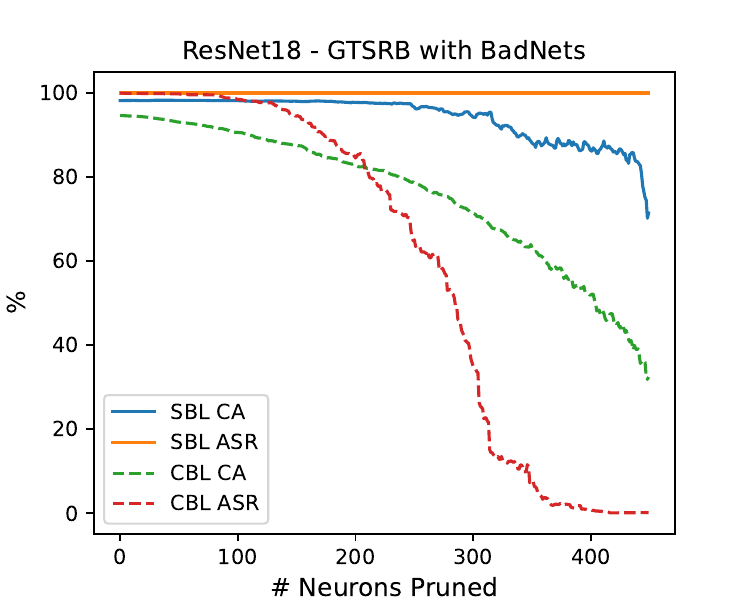}
    % \label{fig:MI_gtsrb_badnet_joint}
  \end{subfigure}

    % \raisebox{12mm}{\rotatebox{90}{\tiny VGG-16}}%
  \begin{subfigure}{0.48\linewidth}
    \centering
    \includegraphics[width=\linewidth]{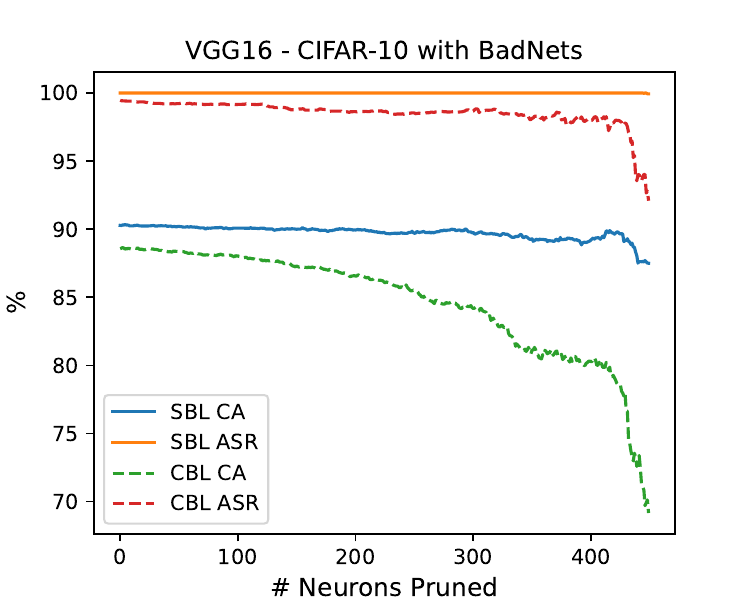}
    % \label{fig:MI_cifar_badnet_sbl_f2c}
  \end{subfigure}
    % \hfill
    \begin{subfigure}{0.48\linewidth}
    \centering
    \includegraphics[width=\linewidth]{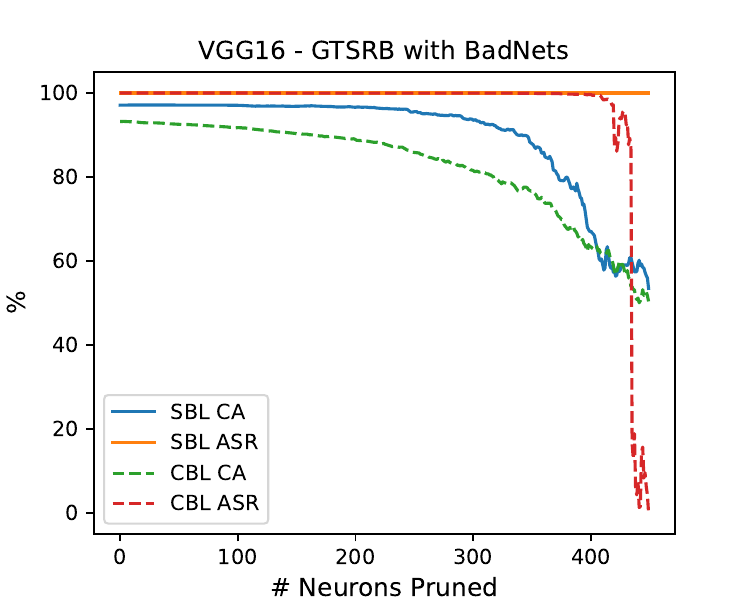}
    % \label{fig:prunin}
  \end{subfigure}

    % \raisebox{12mm}{\rotatebox{90}{\tiny ResNet20}}%
  \begin{subfigure}{0.48\linewidth}
    \centering
    \includegraphics[width=\linewidth]{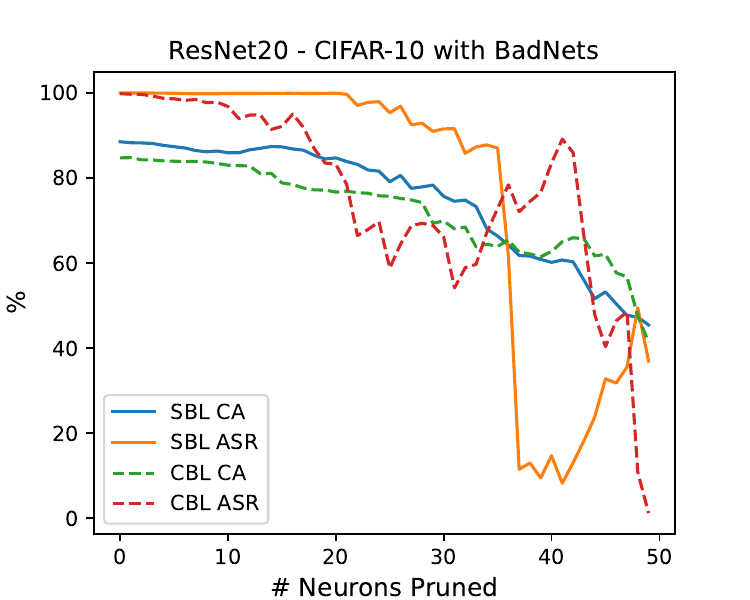}
    % \label{fig:MI_cifar_badnet_sbl}
  \end{subfigure}
    % \hfill
  \begin{subfigure}{0.48\linewidth}
    \centering
    \includegraphics[width=\linewidth]{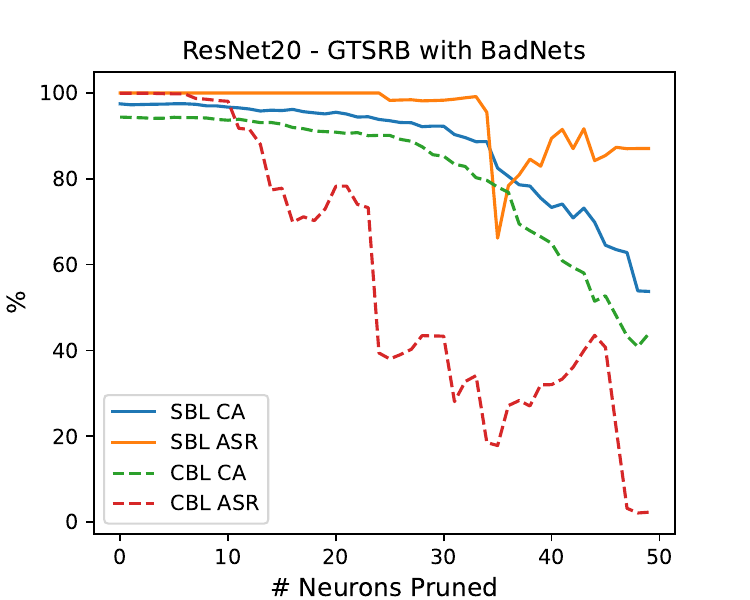}
    % \label{fig:MI_gtsrb_badnet_sbl}
  \end{subfigure}
  
  % \vspace{-0.4cm}

  \caption{Performance of BadNets training with CBL and SBL against Pruning with different settings.}
  \label{fig:pruning}
  % \vspace{-0.5cm}
\end{figure}

Besides fine-tuning defense methods, we also test our methods with the pruning-based approach, in particular, we sequentially prune filters in the last layers of networks. We perform pruning defense on two datasets CIFAR-10 and GTSRB on three different backbones including ResNet18, VGG-16, and ResNet20 with BadNets data poisoning method. In Figure~\ref{fig:pruning}, we visualize the Clean Accuracy (CA) and Attack Success Rate (ASR) of backdoored models trained by Conventional Backdoor Learning (CBL) and our framework Sequential Backdoor Learning (SBL) when pruning each filter or neuron. 
In general, our SBL can help the backdoored model to be more resistant against pruning defense. 
Liu et. al., \cite{liu2018fine} observes that backdoored neurons are dormant for clean inputs, therefore they proposed to prune low-sensitivity neurons to clean data sequentially. Meanwhile, our SBL mitigates this sensitivity through SAM~\cite{foret2021sharpnessaware} which forces the backdoored model's parameters to be more stable with perturbations.

\subsection{Additional Visualizations with Model Interpolation}

\label{sec:sup_ml}
\begin{figure}
  \centering
  % Second row with two figures
  \raisebox{5mm}{\rotatebox{90}{\tiny ResNet20 on CIFAR-10}}%
  \begin{subfigure}{0.48\linewidth}
    \centering
    \includegraphics[width=\linewidth]{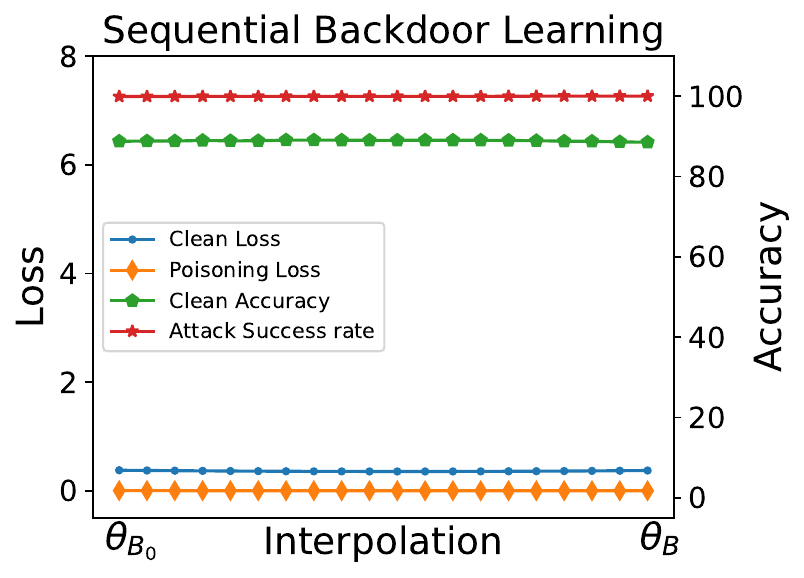}
    % \label{fig:MI_cifar_badnet_joint}
  \end{subfigure}
% \hfill
  \begin{subfigure}{0.48\linewidth}
    \centering
    \includegraphics[width=\linewidth]{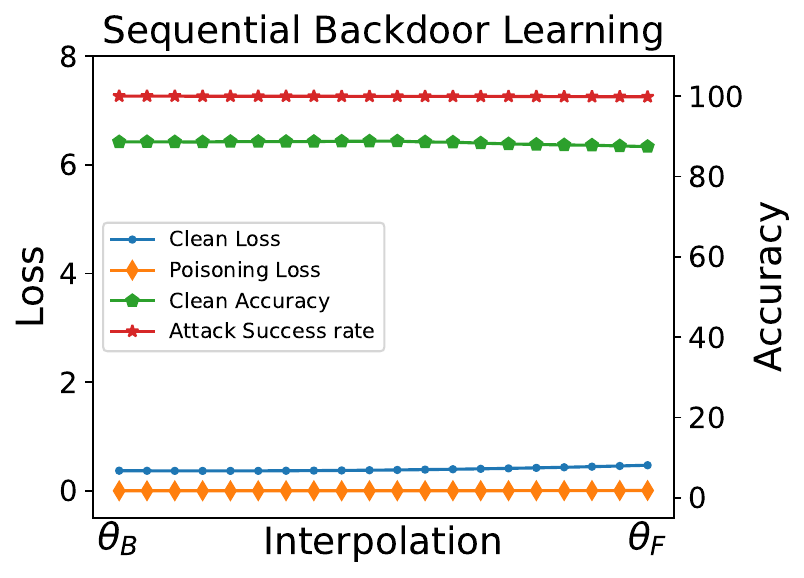}
    % \label{fig:MI_cifar_badnet_sbl_f2c}
  \end{subfigure}

  % Second row with two figures
  \raisebox{5mm}{\rotatebox{90}{\tiny VGG-16 on CIFAR-10}}%
  \begin{subfigure}{0.48\linewidth}
    \centering
    \includegraphics[width=\linewidth]{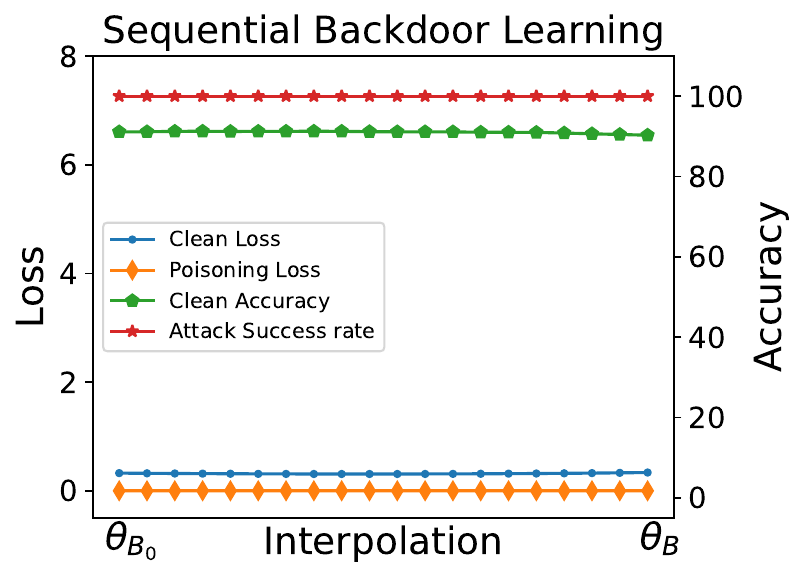}
    % \label{fig:MI_cifar_badnet_joint}
  \end{subfigure}
% \hfill
  \begin{subfigure}{0.48\linewidth}
    \centering
    \includegraphics[width=\linewidth]{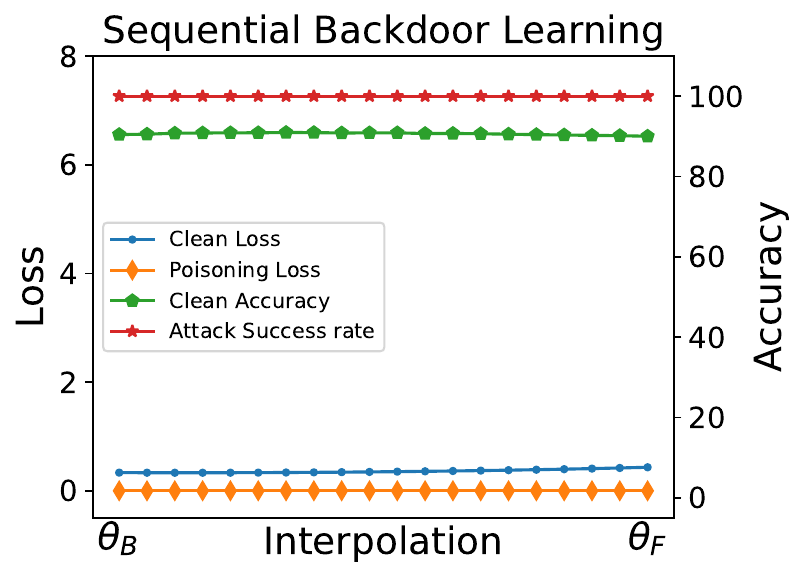}
    % \label{fig:MI_cifar_badnet_sbl_f2c}
  \end{subfigure}

    % Second row with two figures
  \raisebox{5mm}{\rotatebox{90}{\tiny ResNet20 on GTSRB}}%
  \begin{subfigure}{0.48\linewidth}
    \centering
    \includegraphics[width=\linewidth]{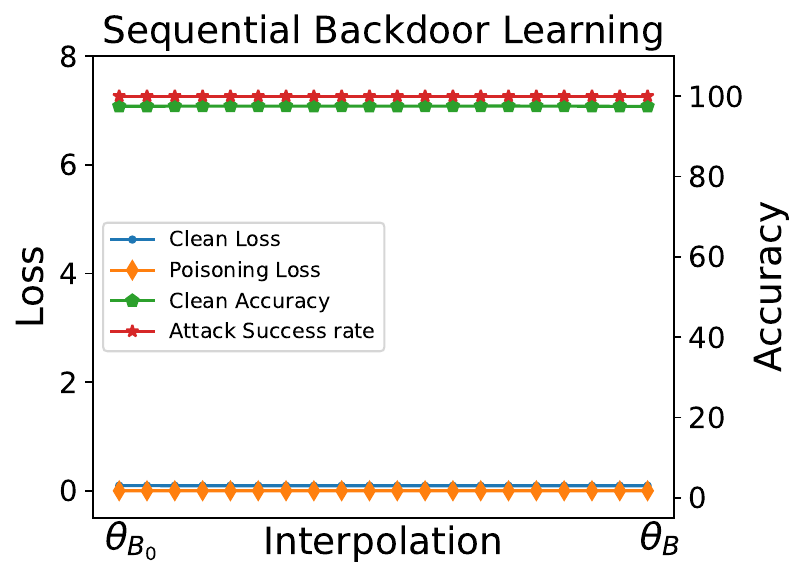}
    % \label{fig:MI_cifar_badnet_joint}
  \end{subfigure}
% \hfill
  \begin{subfigure}{0.48\linewidth}
    \centering
    \includegraphics[width=\linewidth]{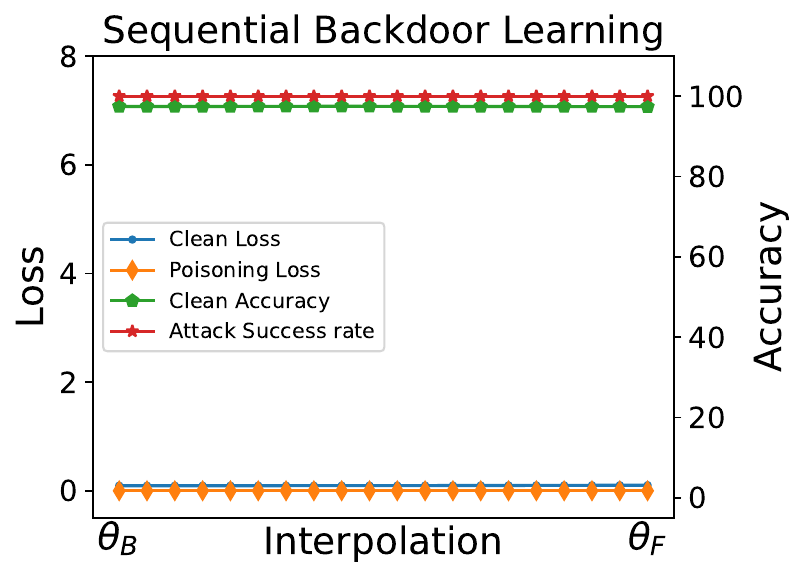}
    % \label{fig:MI_cifar_badnet_sbl_f2c}
  \end{subfigure}

    % Second row with two figures
  \raisebox{5mm}{\rotatebox{90}{\tiny VGG-16 on GTSRB}}%
  \begin{subfigure}{0.48\linewidth}
    \centering
    \includegraphics[width=\linewidth]{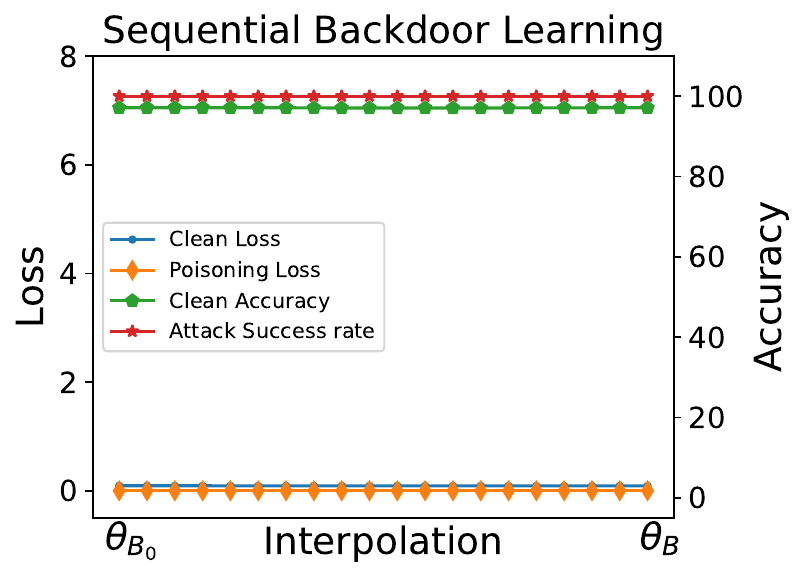}
    % \label{fig:MI_cifar_badnet_joint}
  \end{subfigure}
% \hfill
  \begin{subfigure}{0.48\linewidth}
    \centering
    \includegraphics[width=\linewidth]{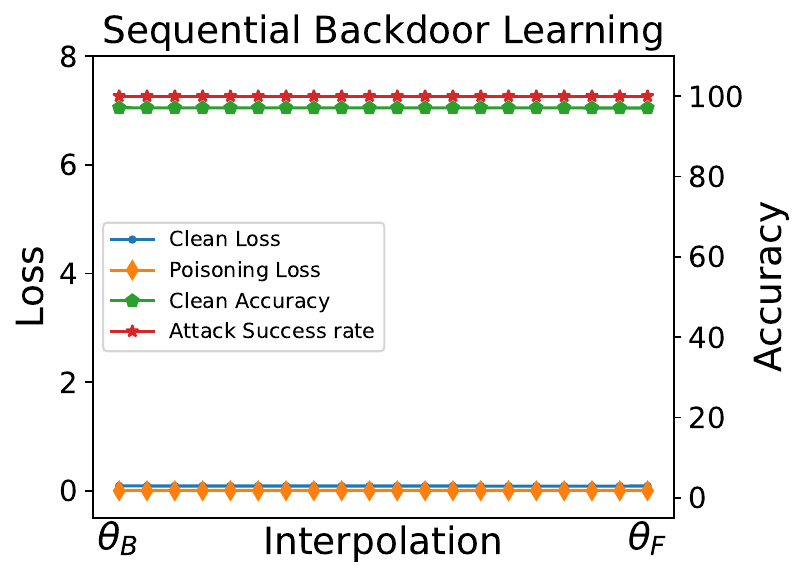}
    % \label{fig:MI_cifar_badnet_sbl_f2c}
  \end{subfigure}
  
  % \vspace{-0.4cm}

  \caption{The loss and the accuracy on clean and poisoned test sets of intermediate models when linearly interpolating between models. The first column is between models in the first ($\theta_{B_0}$) and second task ($\theta_B$), while the second column is between backdoored and fine-tuned models in our SBL framework on different backbones.}
  \label{fig:sup_model_interpolation}
  % \vspace{-0.5cm}
\end{figure}

In this section, we visualize the loss and accuracy on the clean and poisoned test sets while linearly interpolating between model after Step 0 $\theta_{B_0}$ and backdoored model $\theta_B$ in the first column of Figure~\ref{fig:sup_model_interpolation}, and backdoored model $\theta_B$ and fine-tuned model $\theta_F$ on different architecture settings. Figure~\ref{fig:sup_model_interpolation} one more time confirms that our framework SBL can identify a low-error path connecting the backdoored model to the fine-tuned model.

\subsection{Additional Visualizations with Gradient Norm}
\label{sec:sup_gn}
\begin{figure}
% \vspace{-5cm}
  \centering
  % Second row with two figures
  % \raisebox{5mm}{\rotatebox{90}{\tiny }}%
  \begin{subfigure}{0.48\linewidth}
    \centering
    \includegraphics[width=\linewidth]{figures/grad_norm_res/resnet18_cifar10.pdf}
    % \label{fig:MI_cifar_badnet_joint}
  \end{subfigure}
% \hfill
\begin{subfigure}{0.48\linewidth}
    \centering
    \includegraphics[width=\linewidth]{figures/grad_norm_res/resnet18_gtsrb.pdf}
    % \label{fig:MI_gtsrb_badnet_joint}
  \end{subfigure}

  \begin{subfigure}{0.48\linewidth}
    \centering
    \includegraphics[width=\linewidth]{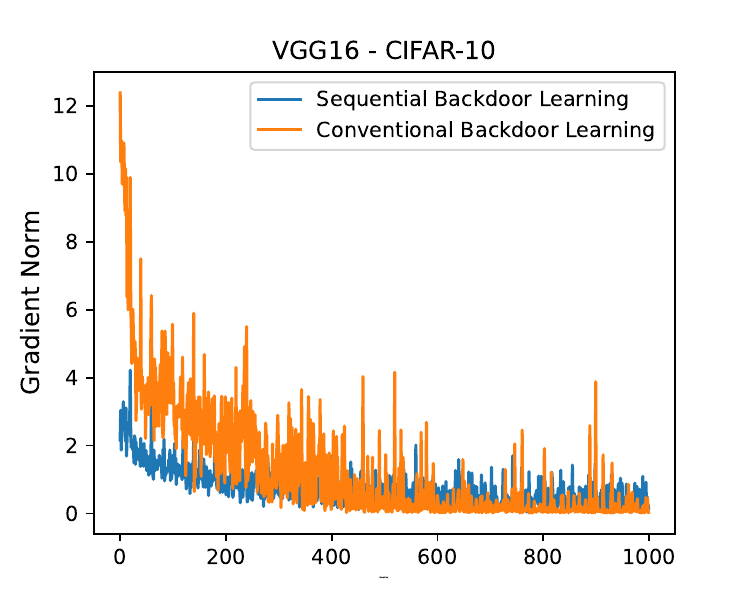}
    % \label{fig:MI_cifar_badnet_sbl_f2c}
  \end{subfigure}
  % \hfill
  \begin{subfigure}{0.48\linewidth}
    \centering
    \includegraphics[width=\linewidth]{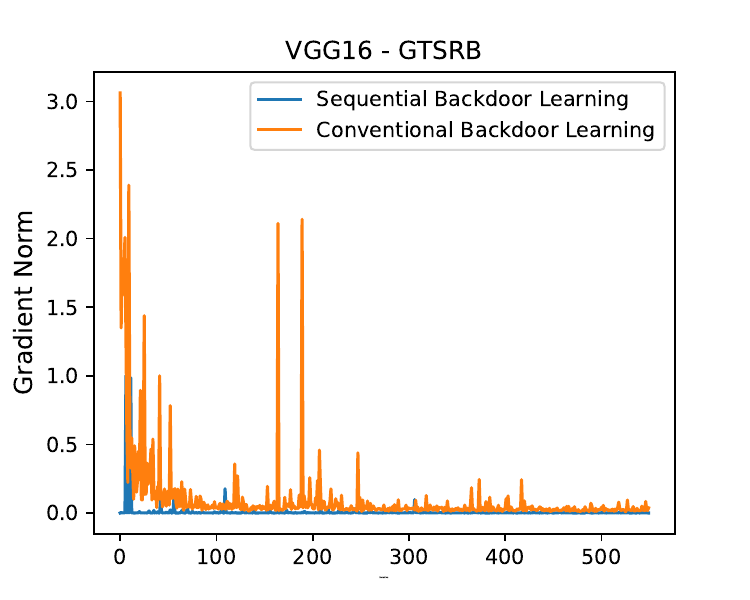}
    % \label{fig:prunin}
  \end{subfigure}

% \vspace{-0.4cm}
% Second row with two figures
    % \raisebox{6mm}{\rotatebox{90}{\tiny ResNet18 on GTSRB}}%

  \begin{subfigure}{0.48\linewidth}
    \centering
    \includegraphics[width=\linewidth]{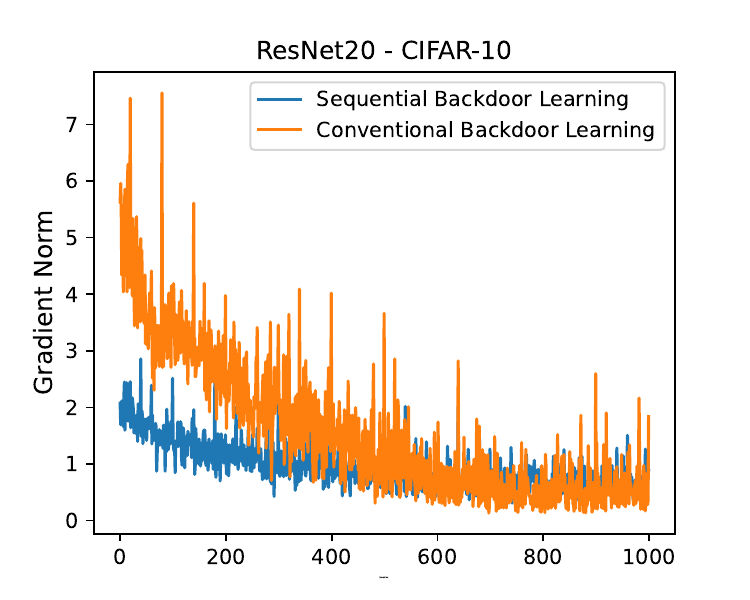}
    % \label{fig:MI_cifar_badnet_sbl}
  \end{subfigure}
  % \hfill
  \begin{subfigure}{0.48\linewidth}
    \centering
    \includegraphics[width=\linewidth]{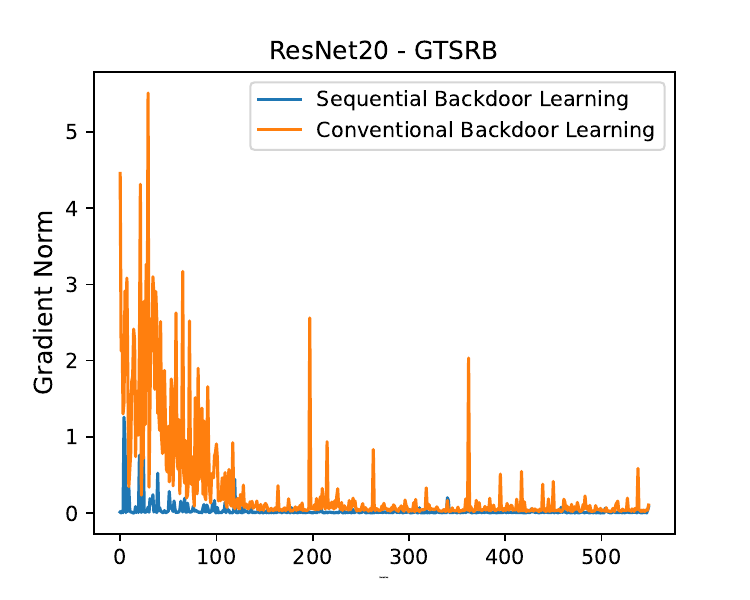}
    % \label{fig:MI_gtsrb_badnet_sbl}
  \end{subfigure}
  
  % \vspace{-0.4cm}

  \caption{Gradient norm comparison between conventional backdoor learning and our framework during fine-tuning with SGD-0.01 from the backdoored model at defense stage on different network architectures.}
  \label{fig:sup_grad_norm}
  % \vspace{-0.5cm}
\end{figure}

Here, we visualize the gradient norm during fine-tuning of our framework SBL compared to CBL in several network architectures and visualize it in Figure~\ref{fig:sup_grad_norm}. With different backbones, we still observe that in the early stage of fine-tuning, the gradient norm values of CBL are substantially higher than SBL which supports CBL poisoned model escape from backdoor regions.

\subsection{Effect of Fine-tuning epochs}
\label{sec:sup_epochs}
\begin{figure}
    \centering
    \includegraphics[width=\linewidth]{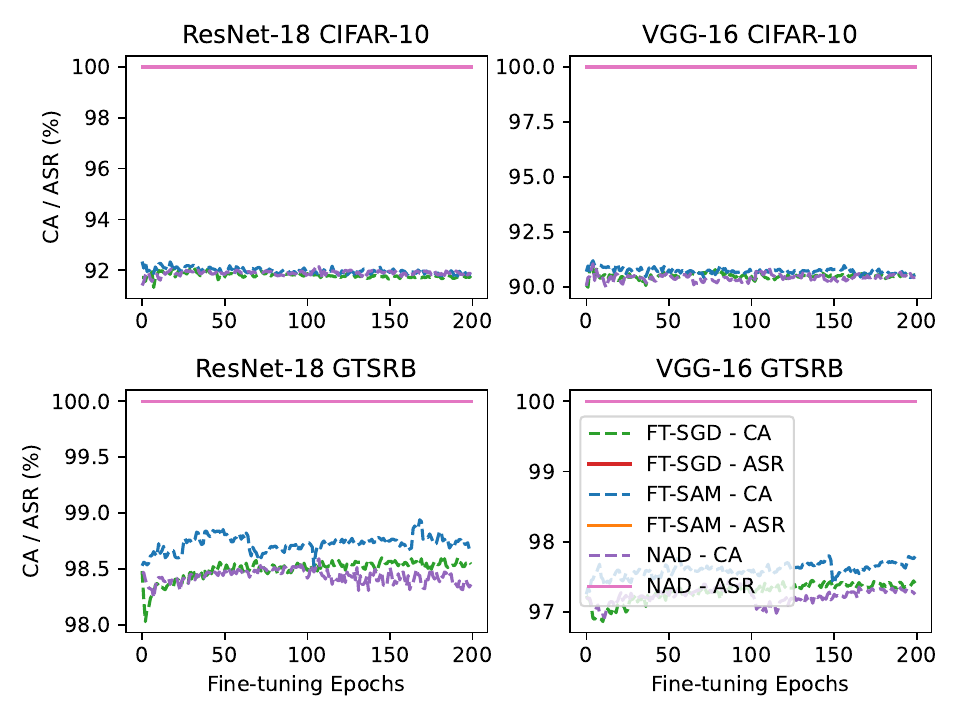}
    \caption{Performance of SBL with more fine-tuning epochs.}
    \label{fig:ft_epochs}
\end{figure}

We investigate the ASR/CA results of SBL against FT defenses for up to 200 epochs in Fig.~\ref{fig:ft_epochs} (BadNet trigger, ResNet-18/VGG-16, CIFAR-10/GTSRB). We can observe that during fine-tuning, the ASR remains the same (100\%) while the CA slightly changes, demonstrating that the model's still trapped in backdoored regions while finding good local minima for clean data.

\section{Limitations}
\label{sec:limitations}
As discussed, this paper aims to improve the resistance of existing backdoor attacks against fine-tuning defenses, thus our empirical evaluation focuses on studying its effectiveness against these tuning defenses. Nevertheless, since our method only augments the existing attacks against fine-tuning defenses, the base attack's effectiveness against other types of defense should be preserved. 

Another limitation of this work is that our SBL requires the adversary to have full access to the training process. While this is a commonly studied backdoor setting in the literature, there exist other settings, e.g., dataset control only, where the adversary cannot control the training processes in practice. Extending SBL framework to perform an attack in one of these settings would be an interesting future extension of our work.

Finally, our work emphasizes the potential risks associated with dependence on pre-trained models from third-party sources, emphasizing the critical need to cultivate trust between users and model providers. To safeguard against these potential threats, users are advised to only utilize pre-trained models from reputable providers or actively involved in the training process. Furthermore, we encourage the research community to launch more investigations into this domain to establish better safeguards.

\end{document}